\documentclass{article}

\usepackage{booktabs,longtable,siunitx,array,makecell}
\usepackage{PRIMEarxiv}
\usepackage{graphicx}
\usepackage[utf8]{inputenc} % allow utf-8 input
\usepackage[T1]{fontenc}    % use 8-bit T1 fonts
\usepackage{hyperref}       % hyperlinks
\usepackage{url}            % simple URL typesetting
\usepackage{booktabs}       % professional-quality tables
\usepackage{amsmath}        % math symbols and environments
\usepackage{amsfonts}       % blackboard math symbols
\usepackage{nicefrac}       % compact symbols for 1/2, etc.
\usepackage{microtype}      % microtypography
\usepackage{lipsum}
\usepackage{fancyhdr}       % header
\usepackage{graphicx}       % graphics
\graphicspath{{media/}}     % organize your images and other figures under media/ folder
\usepackage{setspace}
\usepackage{lineno}

\title{LingLanMedBench: A Unified Benchmark for Evaluating Large Language Models in Traditional Chinese Medicine
}

\author{
  Rui Hua\textsuperscript{†,1},
  Yu Wei\textsuperscript{†,2,3,4},
  Zixin Shu\textsuperscript{†,5,6,7},
  Kai Chang\textsuperscript{1},
  Dengying Yan\textsuperscript{1},
  Jianan Xia\textsuperscript{1},
  Hui Zhu\textsuperscript{5,6,7}, \\
  Xiaodong Li\textsuperscript{5,6,7},
  Dongmei Jia\textsuperscript{8},
  Zhuye Gao\textsuperscript{9},
  Yanyan Meng\textsuperscript{10},
  Haibin Yu\textsuperscript{11},
  Fei Dong\textsuperscript{12},
  Kan Ding\textsuperscript{13}, \\
  Runshun Zhang\textsuperscript{14},
  Jian Yu\textsuperscript{1},
  Jiansheng Li\textsuperscript{*,11},
  Baoyan Liu\textsuperscript{*,15},
  Wenjia Wang\textsuperscript{*,2,3,4},
  Xuezhong Zhou\textsuperscript{*,1}
  \\
  \\
  \textsuperscript{1}School of Computer Science \& Technology, Beijing Jiaotong University, Beijing 100044, China \\
  \textsuperscript{2}Tianjin Tasly Digital Chinese Medicine Technology Co., Ltd., Tianjin, 300410, China \\
  \textsuperscript{3}Tasly Biopharmaceuticals Co., Ltd., Tianjin, 300410, China \\
  \textsuperscript{4}State Key Laboratory of Chinese Medicine Modernization, Tianjin, 300193, China \\
  \textsuperscript{5}Institute of Liver Diseases, Hubei Key Laboratory of the theory and application research of liver and kidney \\
  in traditional Chinese medicine, Hubei Provincial Hospital of Traditional Chinese Medicine, Wuhan, 430061, China \\
  \textsuperscript{6}Affiliated Hospital of Hubei University of Chinese Medicine, Wuhan, 430061, China \\
  \textsuperscript{7}Hubei Province Academy of Traditional Chinese Medicine, Wuhan, 430074, China \\
  \textsuperscript{8}Da'an Health Technology (Beijing) Co., Ltd., Beijing 100000, China \\
  \textsuperscript{9}Xiyuan Hospital, China Academy of Chinese Medical Sciences, Beijing 100091, China \\
  \textsuperscript{10}Beijing Research Institute of Chinese Medicine, Beijing University of Chinese Medicine, Beijing 100029, China \\
  \textsuperscript{11}The First Affiliated Hospital, Henan University of Chinese Medicine, Zhengzhou 450000, China \\
  \textsuperscript{12}Institute of Chinese Medicine Epidemic Disease, Beijing University of Chinese Medicine, Beijing 102488, China \\
  \textsuperscript{13}China Institute for History of Medicine and Medical Literature,\\ China Academy of Chinese Medical Sciences, Beijing 100700, China \\
  \textsuperscript{14}Department of Gastroenterology, Guang'anmen Hospital, \\ China Academy of Chinese Medical Sciences, Beijing 100053, China \\
  \textsuperscript{15}China Academy of Chinese Medical Sciences, Beijing 100700, China \\
  † These authors contributed equally to this work. \\
  * Corresponding author. \\
  Email: li\_js8@163.com, liuby5505@139.com, tsl-wangwenjia@tasly.com, xzzhou@bjtu.edu.cn \\
  Tel: +86-10-51683844
}

\begin{document}
\maketitle
\thispagestyle{fancy}

\doublespacing
\nolinenumbers

\noindent\textbf{Running title:} LingLanMedBench Benchmark for TCM Large Language Models


\clearpage

\begin{abstract}

% Large language models (LLMs) are advancing rapidly in medical natural language processing (NLP), but Traditional Chinese Medicine (TCM) requires domain-faithful evaluation because of its distinctive ontology, terminology, and reasoning patterns. Existing TCM benchmarks remain fragmented in coverage and scale and often use inconsistent or generation-heavy scoring, hindering fair comparison. We present LingLanMedBench, a large-scale, expert-curated, multi-task benchmark that evaluates LLMs across knowledge recall, multi-hop reasoning, information extraction, and clinical decision-making. LingLanMedBench provides unified metrics, synonym-tolerant clinical-label matching, 400-item Hard subsets for each task, and decision-recognition reformulations of diagnosis and treatment tasks. We evaluate 14 leading open-source and proprietary LLMs under a zero-shot protocol. Results reveal strong performance on knowledge-oriented tasks but substantial gaps from human experts on Hard subsets, especially in TCM-specialized reasoning. LingLanMedBench offers a quantitative and extensible foundation for reproducible evaluation of TCM LLMs and domain-specific medical artificial intelligence.

Reliable evaluation of large language models in traditional Chinese medicine remains limited because existing benchmarks mainly focus on examination-style question answering or isolated generative tasks. We developed LingLanMedBench, a unified benchmark comprising 25,620 expert-verified items across 13 subtasks and five domains, covering knowledge, language understanding, information extraction, and diagnostic-therapeutic decision-making. LingLanMedBench standardizes task formats and evaluation metrics and further incorporates synonym-tolerant clinical-label matching, empirically derived hard subsets, and decision-recognition reformulations of open-ended clinical tasks. Fourteen open-source and proprietary models were evaluated under a standardized zero-shot protocol. Performance varied substantially across domains, model families, task formulations, and difficulty levels. The leading overall score decreased from 51.1 on the full benchmark to 31.9 on the hard subsets. Prescription performance also differed markedly between open-ended generation, with F1 scores of 32.7 and 20.3, and decision recognition, with accuracies of 90.6 and 56.8 on the full and hard subsets, respectively. These findings reveal a substantial gap between knowledge-oriented performance and clinically faithful reasoning. LingLanMedBench provides a standardized and reproducible framework for characterizing current capabilities, identifying systematic limitations, and tracking meaningful progress in large language models for traditional Chinese medicine.

\end{abstract}

\clearpage

\section{Introduction}

Foundation models trained on web scale corpora have redefined language understanding and reasoning, with rapid advances across capability and scale~\cite{brown2020language,achiam2023gpt,touvron2023llama,vaswani2017attention}. This shift from task specific systems to general purpose language intelligence has motivated unified evaluations that prioritize breadth and comparability, exemplified by MMLU~\cite{hendrycks2020measuring} and HELM~\cite{liang2022holistic}. 
Contemporary model families such as the GPT family~\cite{achiam2023gpt,agarwal2025gpt}, DeepSeek~\cite{guo2025deepseek}, and Qwen~\cite{yang2025qwen3} illustrate a trajectory toward stronger multistep reasoning and cross domain transfer.

In medicine, frontier large language models (LLMs) can approach or even surpass human performance on standardized knowledge assessments such as MedQA~\cite{jin2021disease}, CMB~\cite{wang2023cmb}, and CMExam~\cite{liu2023benchmarking}, alongside progress in biomedical and clinical modeling (for example, BioGPT and Med-PaLM) and applied evaluation of clinical use cases~\cite{luo2022biogpt,singhal2025toward,liu2025application,moor2023foundation}. Clinical reasoning, however, demands more than factual recall and requires the integration of structured knowledge with contextual patient data and accountable decision processes~\cite{singhal2023large,liu2025evaluating,goh2024large}. Recent surveys therefore emphasize multi-dimensional and clinically grounded evaluation beyond narrow question answering (QA), while noting that most benchmarks remain anchored in modern biomedicine~\cite{tordjman2025comparative}.
Within the Chinese medical ecosystem, general-domain and biomedical LLMs such as Baichuan~\cite{yang2023baichuan}, Baichuan-M2~\cite{dou2025baichuan}, Huatuo~\cite{wang2023huatuo}, and Zhongjing~\cite{yang2024zhongjing} have been adapted to consultation and decision support.
Building on these efforts, Traditional Chinese Medicine (TCM)-oriented models such as TCMChat~\cite{dai2024tcmchat}, Lingdan~\cite{hua2024lingdan}, and BianCang~\cite{wei2025biancang} fine tune or continue training on curated corpora spanning classical canons, syndrome and treatment guidelines, hospital electronic medical records (EMRs), pharmacopoeial knowledge, and prescription data to better reflect TCM's knowledge system and reasoning paradigm.

Across the history of artificial intelligence (AI), benchmark datasets have played a pivotal role in catalyzing progress. 
ImageNet~\cite{deng2009imagenet} established a large-scale standardized testbed for computer vision, while language and reasoning benchmarks such as MMLU~\cite{hendrycks2020measuring}, C-Eval~\cite{huang2023ceval}, HELM~\cite{liang2022holistic}, MATH~\cite{hendrycks2021measuring}, GPQA~\cite{rein2024gpqa}, and Humanity's Last Exam~\cite{phan2025humanity} have extended evaluation toward broad knowledge, formal reasoning, calibration, and expert-level problem solving.
In medicine, MedQA~\cite{jin2021disease}, PubMedQA~\cite{jin2019pubmedqa}, CMB~\cite{wang2023cmb}, CMExam~\cite{liu2023benchmarking}, MedBench~\cite{cai2024medbench}, MedHELM~\cite{bedi2025medhelm}, and Humanity's Next Medical Exam~\cite{gallifant2025humanity} have advanced reproducible assessment of biomedical knowledge and clinical reasoning. 
However, these resources are largely grounded in general or modern biomedical paradigms and provide limited coverage of TCM-specific ontology, terminology, syndrome differentiation, prescription reasoning, and dosage proportionality, thereby limiting the systematic advancement of AI models in this domain.

TCM embodies a deeply empirical and experience-driven medical paradigm.
Unlike the protocolized frameworks of Western biomedicine, TCM emphasizes the recognition of dynamic patterns rather than fixed disease entities, and the corresponding therapeutic principle is determined through interpretive reasoning grounded in accumulated clinical experience rather than strictly codified rules~\cite{lukman2007computational,song2024ai}.
This experience-centered reasoning extends to prescription formulation, where herbal combinations are adjusted according to subtle patient variations, temporal changes, and practitioner intuition.
Such contextual flexibility makes TCM both highly adaptive and inherently difficult to formalize.
From a computational perspective, these characteristics create unique challenges for model evaluation:
the same clinical presentation may correspond to multiple legitimate syndromes or treatment strategies, while similar herbal prescriptions can differ in dosage proportions yet remain therapeutically equivalent.
Moreover, TCM texts often employ metaphorical and non-literal expressions rooted in classical Chinese, which complicate token-level language modeling and semantic normalization.
As a result, an effective benchmark must assess not only factual knowledge retrieval but also models’ ability to approximate the interpretive, experience-based reasoning that underpins authentic TCM practice.

Recent TCM-oriented benchmarks have begun to address this gap. TCMD~\cite{yu2024tcmd} and TCMBench~\cite{yue2024tcmbench} mainly evaluate examination-style TCM knowledge, MTCMB~\cite{kong2025mtcmb} broadens evaluation to multiple task categories, and TCMEval-SDT~\cite{wang2025tcmeval} focuses on syndrome differentiation through staged expert annotation. Despite these advances, existing resources remain limited in several respects: task coverage is often concentrated on question answering or generation, evaluation protocols are not fully harmonized across heterogeneous task formats, and clinically grounded reasoning, including information extraction, syndrome-treatment alignment, prescription composition, and dosage proportionality, remains insufficiently assessed. These limitations motivate a benchmark that combines broad TCM knowledge coverage with reproducible, task-aligned metrics for both structured and reasoning-intensive scenarios.

We introduce LingLanMedBench, a rigorously curated benchmark that elevates evaluation of LLMs for TCM from factual recall to clinically salient reasoning. LingLanMedBench provides broad domain coverage, expert-audited content, hard subsets for robustness, and a unified metric suite that enables commensurate, reproducible comparison across heterogeneous task formats. Under a standardized zero-shot protocol, results show near-ceiling performance on licensing-style recall but persistent deficits in multi-hop synthesis, prescription composition, and dosage proportionality, underscoring the gap between surface knowledge and expert-level TCM reasoning.

Our contributions are as follows: \\
(1) A comprehensive, expert-curated TCM suite spanning five domains and 13 subtasks.\\
(2) A unified, reproducible metric framework with task-aligned scoring and difficulty-calibrated hard subsets.\\
(3) Systematic cross-model baselines that expose actionable failure modes in option calibration, information extraction scaling, and clinical reasoning.

\section{Materials and methods}

We now detail LingLanMedBench's construction, covering its data sources and curation, task taxonomy, metric definitions, and standardized zero-shot evaluation settings, to provide a reproducible blueprint for future extensions.

\subsection{Overview}

\begin{figure}[!t]
  \centering
  \includegraphics[width=0.95\textwidth]{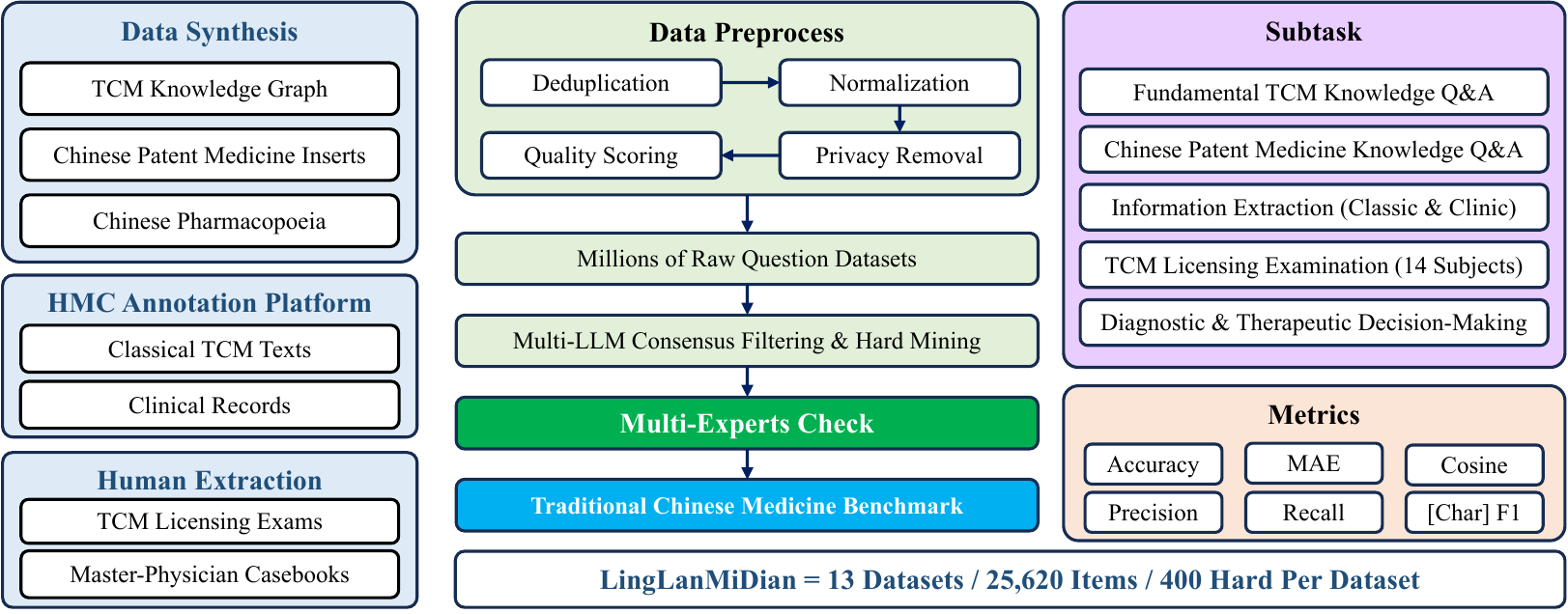}
  \caption{Overview of the LingLanMedBench construction pipeline. The left panel summarizes data sources; the center panel illustrates data processing and curation; the right panel presents the task taxonomy covered in LingLanMedBench together with the unified evaluation metrics. Abbreviations: HMC (Human-machine collaboration).}
  \label{fig:overview}
\end{figure}

LingLanMedBench benchmark is designed to systematically evaluate the performance of LLMs across the full spectrum of TCM knowledge and reasoning. 
It unifies heterogeneous evaluation tasks such as knowledge recall, multi-hop reasoning, information extraction, and diagnostic decision-making under a consistent and quantitative framework, as illustrated in Figure~\ref{fig:overview}.
LingLanMedBench comprises 13 datasets spanning five domains: \textit{Licensing Examination}, \textit{Fundamental TCM Knowledge}, \textit{Chinese Patent Medicine}, \textit{Information Extraction}, and \textit{Diagnostic and Therapeutic Decision-Making}. 
In total, the benchmark contains 25,620 expert-verified items, with approximately 2,000 samples per dataset. For each dataset, a 400-item \textit{Hard subset} is selected from the full set to assess advanced reasoning capability.

\subsection{Data Sources}

\textbf{(1) TCM Licensing Examination (TLE).} 
Questions were collected from the official National TCM Qualification Examination (2000–2016), covering a comprehensive range of medical subjects including \textit{Fundamentals of TCM Theory}, \textit{Diagnostics of TCM}, \textit{Chinese Materia Medica}, \textit{Formulary Science}, \textit{Internal Medicine of TCM}, \textit{Surgery of TCM}, \textit{Gynecology of TCM}, \textit{Pediatrics of TCM}, and \textit{Acupuncture and Moxibustion}. 
In addition, the dataset integrates supporting courses required in medical licensing education, such as \textit{Fundamentals of Diagnostics}, \textit{Internal Medicine}, \textit{Infectious Diseases}, \textit{Medical Ethics}, and \textit{Health Regulations}. 
Together, these subjects comprehensively represent the standardized knowledge framework assessed in national medical education and form the foundation for evaluating both factual and procedural competence in TCM.

\textbf{(2) Fundamental TCM Knowledge (FTK).}
This dataset is grounded in a self-constructed TCM knowledge graph~\cite{hua2026lingshu} assembled from authoritative sources, including canonical classical texts, contemporary peer-reviewed TCM/biomedical literature, and nationally adopted teaching materials, and further augmented with public databases such as SymMap~\cite{wu2019symmap} and TCMID~\cite{huang2018tcmid}. The graph consolidates and normalizes entities and relations across herbs, ingredients, efficacies, symptoms, syndromes, prescriptions, symptom clusters, diseases, and genes, capturing hierarchical and therapeutic links (e.g., herb–ingredient, herb–efficacy, symptom–syndrome, syndrome–prescription) under a unified ontology.

\textbf{(3) Chinese Patent Medicine Knowledge (CPMK).}
This dataset is built from structured knowledge extracted from the Pharmacopoeia of the People’s Republic of China and package inserts of Chinese patent medicines. The source corpus was normalized into a unified schema covering herbal composition, associated diseases and syndromes, pharmacology and toxicology, indications, contraindications, product specifications, safety information, etc. In total, it comprises standardized records for more than 9,000 Chinese patent medicines.

\textbf{(4) Information Extraction (IE).} 
The information extraction datasets draw from two sources: classical TCM literature written in Literary Chinese and real-world clinical medical records composed in modern vernacular Chinese. Annotated samples were produced via a semi-automated pipeline that combines named entity recognition (NER) algorithms with expert verification \cite{zou2022phenonizer}. All clinical texts were fully de-identified prior to annotation to ensure privacy compliance. The resulting corpora encompass more than one hundred entity categories, including positive and negative symptoms, herbs, positive and negative physical signs, condition changes, and episode characteristics. This dual-register design captures both historical and contemporary linguistic styles, enabling fine-grained evaluation of LLMs in structured information extraction and clinical text understanding across both classical and modern contexts.

\textbf{(5) Diagnostic and Therapeutic Decision-Making.}  
The diagnostic and therapeutic datasets are built from more than 13,000 clinical case records authored by over 300 renowned TCM physicians, covering 7,595 syndromes. Domain experts extracted and verified high-quality annotations for syndrome differentiation, treatment principles, herbal prescriptions, and corresponding dosages. We further assessed coverage against the ICD-11 \cite{harrison2021icd} “Traditional medicine patterns” category: excluding “specified” and “unspecified” terms, there are 216 categories in total, of which the dataset covers 174, yielding a coverage rate of 80.56\%. These resources enable holistic evaluation of LLMs on integrated clinical reasoning and decision-making in TCM.

Except for TLE, the benchmark items had not appeared in public internet corpora before model evaluation.

\begin{figure}[htbp]
  \centering
  \includegraphics[width=0.8\textwidth]{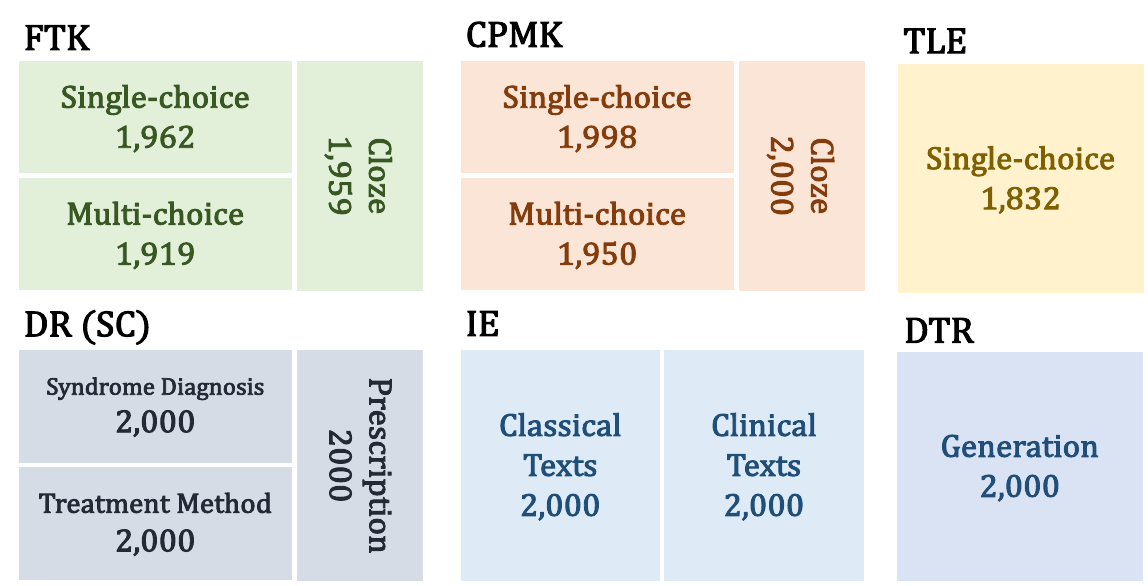}
  \caption{Subtask distribution and dataset sizes in LingLanMedBench. Abbreviations: FTK (Fundamental TCM Knowledge), CPMK (Chinese Patent Medicine Knowledge), TLE (TCM Licensing Examination), DR (Decision Recognition), IE (Information Extraction), DTR (Diagnostic–Therapeutic Reasoning), SC (single-choice).}
  \label{fig:dataset_statistic}
\end{figure}

\subsection{Data Curation and Quality Assurance}

LingLanMedBench comprises 13 subtasks across five domains, evaluated using 44 indicators (Figure~\ref{fig:dataset_statistic}). Data underwent text normalization, de-duplication, automatic screening, and expert verification. Clinical texts were de-identified before annotation. During expert verification, reviewers consulted authoritative references without using LLMs or other automated assistants, corrected ambiguous questions and terminology, verified reference answers, and excluded irreparable items.

\textbf{Knowledge-oriented tasks.}
TLE contains 1,832 single-choice questions with unique reference answers across 14 subjects (Supplementary Fig.~S2). FTK and CPMK questions were generated using Qwen3-32B from knowledge-graph triples and structured medicine records, respectively, in single-choice, multiple-choice, and cloze formats. Automatic screening followed by review by licensed TCM practitioners yielded 5,840 FTK items and 5,948 CPMK items.

\textbf{Information extraction.}
Automatic NER suggestions followed by expert adjudication produced two annotated corpora, each containing 2,000 samples: modern clinical records and classical TCM literature. Together, they cover 106 entity types; their entity distributions are shown in Supplementary Fig.~S3.

\textbf{Diagnostic and therapeutic decision-making.}
DTR comprises 2,000 cases selected from over 13,000 records, requiring complete four-diagnoses notes and prescriptions containing at least five herbs. It evaluates syndrome differentiation, treatment principles, prescriptions, and dosages. DR reformulates each case into three single-choice items addressing syndromes, treatment principles, and prescriptions, each with one correct option and four distractors selected through embedding-based retrieval.

\textbf{Hard subset construction.}
Item difficulty was estimated from scores across 14 models as \(D=(1-\mu)+0.5\sigma\), where \(\mu\) and \(\sigma\) denote the across-model mean and variance, respectively. The 400 highest-scoring items from each subtask formed LingLanMedBench-Hard, totaling 5,200 items (Supplementary Fig.~S1). 

Detailed annotation rules, question-generation procedures, and distractor construction are provided in Supplementary Section~S5.

\subsection{Evaluation Metrics}

Single-choice tasks were evaluated by accuracy. Multiple-choice tasks used exact-set accuracy, requiring all correct options and no incorrect options, together with option-set precision, recall, and F1. Cloze responses were evaluated using character-level F1 based on character multiset overlap.

For information extraction, precision, recall, and F1 were computed from multisets of entity-type and normalized-surface-form pairs, requiring exact agreement of both components. For DTR, predicted and reference labels were normalized, deduplicated, and aligned one-to-one using bidirectional substring inclusion. DTR-F1 allowed synonym-tolerant matching through maximum-cardinality bipartite matching, with an edge defined by character-level F1 \(\geq 0.7\).

Dosage estimation used mean absolute error (MAE) over matched herb pairs and cosine similarity between aligned dosage vectors, with zeros assigned to unmatched herbs. MAE was set to zero when no herb pairs matched; this convention does not indicate perfect dosage accuracy. Cosine similarity was set to one when both vectors had zero norm and zero when only one did.

Scores were macro-averaged across instances within each subtask and reported for the full datasets and Hard subsets. Complete formulas, task-specific matching definitions, and dosage-alignment rules are provided in Supplementary Section~S6.

\subsection{Baseline Models and Evaluation Settings}

We evaluated 14 LLMs spanning general-purpose and medical models, with dense and mixture-of-experts architectures. The Qwen3 models comprised Qwen3-4B, Qwen3-8B, Qwen3-14B, Qwen3-32B, Qwen3-30B-A3B, and Qwen3-235B-A22B~\cite{yang2025qwen3}, together with Qwen3-Next-80B-A3B-Thinking. We also evaluated DeepSeek-R1~\cite{guo2025deepseek}, DeepSeek-V3.1-Think, and the medical model Baichuan-M2-32B~\cite{dou2025baichuan}. The GPT models included GPT-5, GPT-5-mini, and the open-weight models GPT-OSS-20B and GPT-OSS-120B~\cite{agarwal2025gpt}.
Additional Chinese medical and TCM models~\cite{dai2024tcmchat,hua2024lingdan,wei2025biancang} were evaluated for comparison, with results reported in Supplementary Table~S1. Further descriptions of the baseline models are provided in Supplementary Section~S4.

All models were evaluated under a zero-shot setting without any task-specific prompt engineering or fine-tuning.
When supported, decoding hyperparameters were kept consistent: temperature T = 0.6, a maximum generation length of 8,192 tokens, and reasoning/thinking mode enabled during inference.
For GPT models, we set \texttt{reasoning\_effort="medium"} to prevent frequent exceedance of the 8,192-token budget observed under the \texttt{"high"} setting, while maintaining sufficient reasoning depth.

\section{Results}

\begin{figure}
  \centering
  \includegraphics[width=\textwidth]{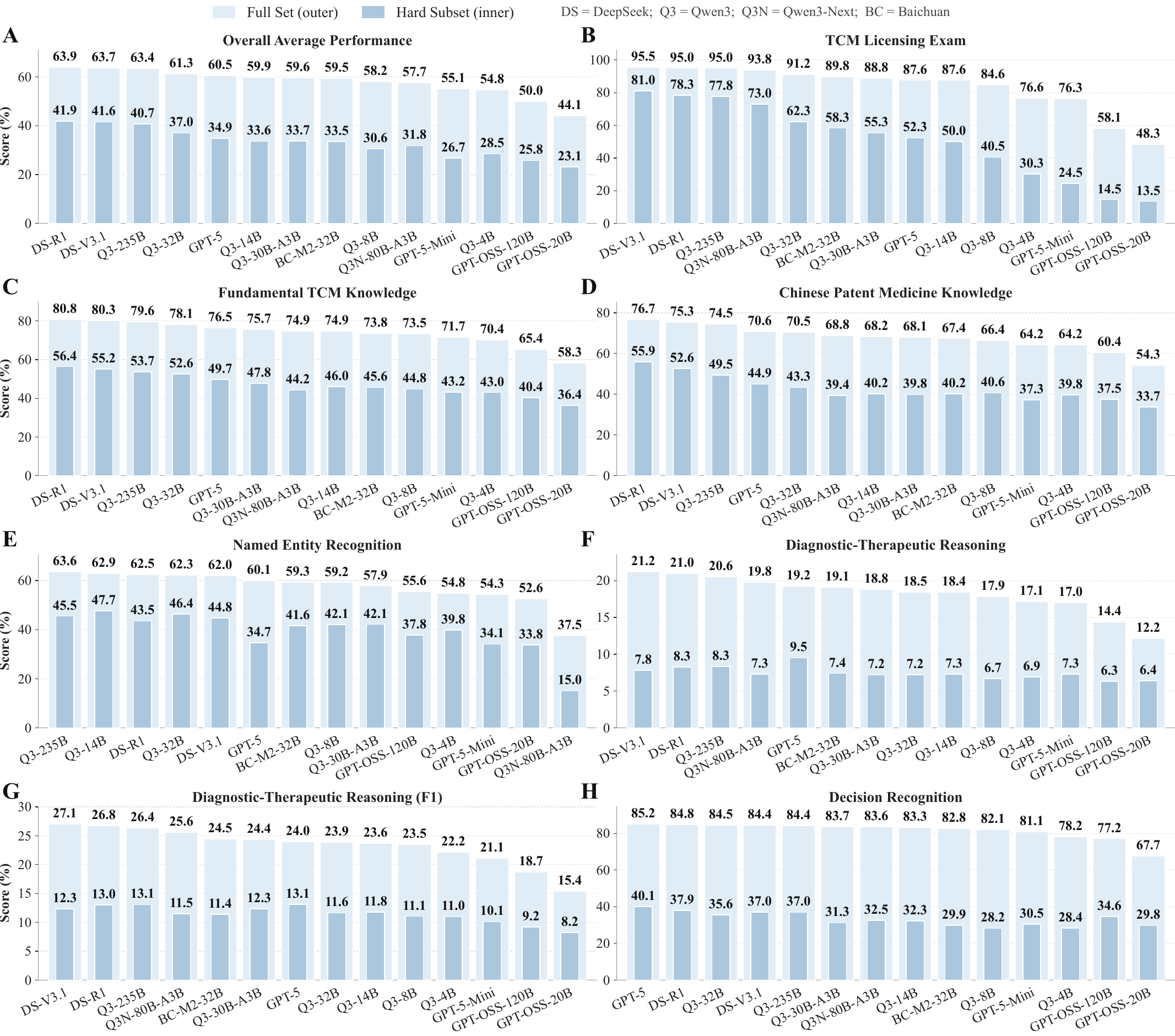}
  \caption{The Overall Average panel shows the mean of the seven task-type scores. The \textit{Average} rows in Tables~\ref{tab:Large-LLM-performance} and~\ref{tab:Small-LLM-performance} instead show the mean of all 42 eligible metric rows.}
  \label{fig:model_performance}
\end{figure}

Under a unified zero-shot protocol, with identical decoding settings where supported, we evaluated 14 models on 13 subtasks spanning five domains: TCM Licensing Examination (TLE), Fundamental TCM Knowledge (FTK), Chinese Patent Medicine Knowledge (CPMK), information extraction (IE), and diagnostic and therapeutic decision-making. The last domain comprises diagnostic and therapeutic reasoning (DTR) and decision recognition (DR). Results are reported for both the full test sets and the curated 400-item hard subsets, and all scores are macro-averaged within each subtask. Information extraction is evaluated as named entity recognition (NER), cloze performance is reported with character-level F1 (char-F1), the synonym-tolerant variant of DTR is denoted DTR-F1, and dosage error is reported using mean absolute error (MAE) together with cosine similarity. The complete results are presented in Tables~\ref{tab:Large-LLM-performance} and~\ref{tab:Small-LLM-performance}, with Qwen3-32B included in both tables as a shared reference model to facilitate comparison.
Overall averages on the full sets cluster in the high 40s to low 50s, while the hard subsets lower the averages to the low 20s to low 30s, indicating a consistent degradation under increased difficulty.
Figure~\ref{fig:model_performance} provides a more intuitive visualization of the performance differences across models on both the full and hard subsets.
The overall average for each model is obtained by first averaging over sub-tasks and metrics (accuracy, F1, cosine, precision, recall) within each task type, then taking the mean of these per-task-type scores so that each task type is weighted equally.

\begin{table*}[t]
  \centering
  \scriptsize
  \setlength{\tabcolsep}{3pt}
  \caption{LingLanMedBench results covering all subtasks. Each entry reports full-set / Hard-subset performance (left/right). Bold values denote the highest score for each subtask–metric. \textit{Average} refers to the mean of all metrics, excluding MAE.}

  \label{tab:Large-LLM-performance}
  \resizebox{\textwidth}{!}{
  \begin{tabular}{lllcccccccc}
    \toprule
    \textbf{Dom.} & \textbf{Subtask} & \textbf{Metric} & \textbf{DeepSeek-R1} & \textbf{DeepSeek-V3.1} & \textbf{GPT-5} & \textbf{Qwen3-235B} & \textbf{Qwen3-32B} & \textbf{Baichuan-M2} & \textbf{Qwen3-Next} & \textbf{GPT-OSS-120B} \\
    \midrule
    % --- TLE
    TLE & Comprehensive & Accuracy & 95.0 / 78.3 & \textbf{95.5} / \textbf{81.0} & 87.6 / 52.3 & 95.0 / 77.8 & 91.2 / 62.3 & 89.8 / 58.3 & 93.8 / 73.0 & 58.1 / 14.5 \\
    % \midrule
    % --- FTK
    FTK & Single-choice & Accuracy & \textbf{86.1} / \textbf{46.5} & \textbf{86.1} / 45.3 & 85.4 / 44.5 & 85.0 / 45.3 & \textbf{86.1} / 44.3 & 83.4 / 35.8 & 82.1 / 33.3 & 70.4 / 28.2 \\
    FTK & Multiple-choice & Accuracy & \textbf{64.8} / \textbf{31.5} & 62.4 / 28.0 & 53.9 / 19.3 & 61.3 / 25.0 & 56.3 / 23.3 & 48.6 / 14.0 & 52.6 / 14.8 & 35.3 / 9.8 \\
    FTK & Multiple-choice & Precision & 92.5 / \textbf{82.4} & 91.9 / 80.5 & 90.6 / 75.3 & \textbf{93.0} / 80.9 & 90.3 / 77.5 & 90.5 / 75.4 & 89.4 / 71.6 & 81.2 / 63.5 \\
    FTK & Multiple-choice & Recall & \textbf{92.1} / 79.3 & 92.0 / \textbf{80.5} & 87.2 / 69.8 & 89.1 / 73.7 & 90.9 / 78.0 & 84.5 / 66.1 & 84.7 / 65.2 & 83.2 / 65.6 \\
    FTK & Multiple-choice & F1 & \textbf{91.0} / \textbf{77.8} & 90.6 / 77.5 & 87.3 / 69.3 & 89.5 / 73.7 & 89.1 / 74.8 & 85.3 / 66.4 & 84.7 / 64.2 & 80.0 / 61.3 \\
    FTK & Cloze & char-F1 & 58.2 / 20.9 & 58.8 / 19.7 & 54.9 / 19.8 & \textbf{59.9} / \textbf{23.5} & 56.1 / 17.5 & 50.2 / 16.2 & 56.0 / 15.9 & 42.1 / 14.0 \\
    % \midrule
    % --- CPMK
    CPMK & Single-choice & Accuracy & \textbf{74.3} / \textbf{31.0} & 73.3 / 25.5 & 67.3 / 18.0 & 68.6 / 17.8 & 66.9 / 16.3 & 60.5 / 12.8 & 63.1 / 8.0 & 44.5 / 13.0 \\
    CPMK & Multiple-choice & Accuracy & \textbf{48.0} / \textbf{24.5} & 42.2 / 15.0 & 35.4 / 6.5 & 44.8 / 15.3 & 35.5 / 6.5 & 33.2 / 5.5 & 34.2 / 3.8 & 24.7 / 4.0 \\
    CPMK & Multiple-choice & Precision & \textbf{86.4} / \textbf{73.5} & 84.3 / 69.6 & 82.4 / 62.7 & 86.1 / 69.7 & 81.9 / 62.1 & 83.1 / 63.5 & 80.6 / 58.7 & 76.0 / 53.9 \\
    CPMK & Multiple-choice & Recall & 90.6 / 76.3 & \textbf{91.5} / \textbf{78.4} & 87.8 / 72.2 & 87.7 / 71.1 & 87.8 / 70.0 & 82.2 / 59.9 & 84.1 / 62.0 & 82.5 / 61.8 \\
    CPMK & Multiple-choice & F1 & \textbf{87.0} / \textbf{72.4} & 86.3 / 71.5 & 83.3 / 64.6 & 85.4 / 68.0 & 83.2 / 63.8 & 80.6 / 58.6 & 80.1 / 57.0 & 77.3 / 55.6 \\
    CPMK & Cloze & char-F1 & 74.0 / \textbf{57.6} & \textbf{74.4} / 55.7 & 67.6 / 45.7 & 74.3 / 54.9 & 67.5 / 41.3 & 64.6 / 41.0 & 70.8 / 46.7 & 57.5 / 36.6 \\
    % \midrule
    % --- NER
    NER & Clinical EMR & Precision & 66.6 / 46.7 & 64.6 / 46.7 & 72.8 / 56.2 & 69.1 / 50.5 & \textbf{73.3} / \textbf{61.3} & 61.8 / 45.5 & 38.1 / 17.3 & 59.7 / 41.4 \\
    NER & Clinical EMR & Recall & 61.3 / 43.9 & 58.6 / 42.4 & \textbf{71.9} / \textbf{56.0} & 66.3 / 49.3 & 62.2 / 49.8 & 56.0 / 41.9 & 33.5 / 15.0 & 55.8 / 41.9 \\
    NER & Clinical EMR & F1 & 63.1 / 44.2 & 60.8 / 43.4 & \textbf{71.7} / \textbf{55.1} & 67.2 / 49.1 & 66.7 / 54.2 & 58.1 / 42.6 & 35.2 / 15.5 & 57.1 / 40.6 \\
    NER & Classical Texts & Precision & 60.1 / 40.2 & \textbf{61.4} / \textbf{42.8} & 44.7 / 11.4 & 57.3 / 37.9 & 57.9 / 38.1 & 58.9 / 38.0 & 38.7 / 13.8 & 51.1 / 31.0 \\
    NER & Classical Texts & Recall & 62.9 / 44.7 & \textbf{64.4} / \textbf{48.8} & 51.7 / 16.3 & 62.6 / 45.9 & 56.9 / 37.9 & 61.6 / 42.6 & 40.3 / 14.9 & 56.6 / 38.5 \\
    NER & Classical Texts & F1 & 60.8 / 41.5 & \textbf{62.1} / \textbf{44.7} & 47.5 / 13.1 & 59.1 / 40.4 & 56.6 / 36.9 & 59.5 / 39.2 & 39.0 / 13.8 & 53.0 / 33.5 \\
    % \midrule
    % --- DTR
    DTR & Syndrome & Precision & 9.9 / 1.3 & \textbf{11.0} / \textbf{1.5} & 5.0 / 0.7 & 9.0 / 1.2 & 7.7 / 0.6 & 9.1 / 1.2 & 10.7 / \textbf{1.5} & 3.2 / 0.6 \\
    DTR & Syndrome & Recall & 13.3 / 1.7 & \textbf{14.3} / 1.8 & 8.4 / 1.1 & 13.4 / \textbf{1.9} & 11.2 / 0.9 & 12.6 / 1.6 & 12.6 / 1.7 & 4.5 / 0.7 \\
    DTR & Syndrome & F1 & 10.9 / 1.5 & \textbf{11.9} / \textbf{1.6} & 6.0 / 0.9 & 10.4 / 1.4 & 8.8 / 0.7 & 10.1 / 1.4 & 11.0 / \textbf{1.6} & 3.6 / 0.6 \\
    DTR & Treatment & Precision & 14.2 / 1.9 & \textbf{14.3} / 0.8 & 10.5 / \textbf{2.1} & 12.9 / 0.9 & 11.5 / 1.3 & 11.7 / 0.6 & 13.6 / 0.6 & 9.6 / 0.9 \\
    DTR & Treatment & Recall & 17.0 / 2.3 & 16.4 / 0.8 & \textbf{19.2} / \textbf{4.6} & 15.6 / 1.2 & 14.4 / 1.5 & 15.0 / 0.9 & 14.5 / 0.9 & 11.2 / 1.2 \\
    DTR & Treatment & F1 & \textbf{14.8} / 1.9 & 14.7 / 0.8 & 13.1 / \textbf{2.8} & 13.6 / 1.0 & 12.3 / 1.3 & 12.6 / 0.7 & 13.5 / 0.7 & 9.9 / 1.0 \\
    DTR & Prescription & Precision & 35.7 / 19.7 & \textbf{36.9} / \textbf{19.9} & 27.6 / 17.1 & 33.4 / 19.0 & 31.0 / 17.0 & 32.1 / 18.0 & 32.4 / 17.4 & 30.2 / 17.3 \\
    DTR & Prescription & Recall & 30.6 / 17.9 & 29.6 / 17.0 & \textbf{39.3} / \textbf{25.9} & 33.5 / 20.1 & 29.3 / 16.9 & 29.4 / 17.2 & 29.9 / 16.8 & 22.9 / 13.6 \\
    DTR & Prescription & F1 & 32.3 / 18.2 & 32.1 / 17.7 & 31.9 / \textbf{20.1} & \textbf{32.7} / 18.9 & 29.5 / 16.4 & 30.0 / 17.0 & 30.4 / 16.4 & 25.4 / 14.7 \\
    DTR & Dosage & MAE & 4.1 / 4.0 & 4.2 / 4.1 & 4.5 / 4.9 & 4.2 / 4.2 & 4.4 / 4.7 & 4.2 / 4.1 & \textbf{4.0} / 3.9 & 4.4 / \textbf{3.7} \\
    DTR & Dosage & Cosine & 30.9 / 16.2 & 31.0 / 16.4 & \textbf{31.4} / \textbf{19.7} & 31.2 / 17.7 & 29.0 / 15.5 & 28.7 / 15.4 & 29.1 / 15.2 & 23.3 / 12.8 \\
    % \midrule
    % --- DTR-F1
    DTR-F1 & Syndrome & Precision & 20.4 / 8.9 & \textbf{21.5} / \textbf{10.0} & 12.1 / 5.8 & 19.0 / 9.5 & 16.5 / 7.8 & 18.1 / 7.9 & 20.9 / 9.3 & 10.6 / 5.4 \\
    DTR-F1 & Syndrome & Recall & 26.6 / 11.2 & 27.1 / 11.2 & 20.1 / 9.1 & \textbf{27.4} / \textbf{12.3} & 23.8 / 10.5 & 24.8 / 10.3 & 24.1 / 9.7 & 14.9 / 6.8 \\
    DTR-F1 & Syndrome & F1 & 22.0 / 9.6 & \textbf{22.9} / 10.1 & 14.6 / 6.9 & 21.6 / \textbf{10.4} & 18.7 / 8.7 & 20.0 / 8.5 & 21.3 / 9.1 & 11.9 / 5.8 \\
    DTR-F1 & Treatment & Precision & 21.0 / \textbf{8.1} & \textbf{21.7} / 6.2 & 15.0 / 5.7 & 19.8 / 6.4 & 18.0 / 6.7 & 18.0 / 5.1 & 21.1 / 5.9 & 14.6 / 4.3 \\
    DTR-F1 & Treatment & Recall & 25.2 / 10.6 & 24.9 / 7.8 & \textbf{28.1} / \textbf{12.3} & 24.1 / 8.7 & 23.0 / 9.0 & 23.3 / 7.3 & 22.5 / 7.2 & 17.7 / 5.9 \\
    DTR-F1 & Treatment & F1 & 22.1 / \textbf{8.9} & \textbf{22.5} / 6.7 & 18.9 / 7.6 & 21.0 / 7.2 & 19.5 / 7.5 & 19.6 / 5.9 & 21.1 / 6.4 & 15.3 / 4.9 \\
    DTR-F1 & Prescription & Precision & 35.6 / 19.8 & \textbf{37.0} / \textbf{20.0} & 27.7 / 17.3 & 33.4 / 19.2 & 30.9 / 17.0 & 31.6 / 18.0 & 33.5 / 17.9 & 30.1 / 17.4 \\
    DTR-F1 & Prescription & Recall & 31.0 / 18.2 & 29.6 / 17.1 & \textbf{39.7} / \textbf{26.1} & 33.6 / 20.2 & 29.6 / 17.0 & 30.2 / 17.8 & 30.8 / 17.2 & 23.0 / 13.7 \\
    DTR-F1 & Prescription & F1 & 32.4 / 18.4 & 32.2 / 17.8 & 32.0 / \textbf{20.3} & \textbf{32.7} / 19.0 & 29.6 / 16.5 & 30.2 / 17.3 & 31.2 / 16.9 & 25.5 / 14.8 \\
    DTR-F1 & Dosage & MAE & 4.1 / 4.0 & 4.2 / 4.2 & 4.5 / 5.0 & 4.2 / 4.2 & 4.4 / 4.7 & 4.2 / 4.2 & \textbf{4.0} / 3.9 & 4.4 / \textbf{3.8} \\
    DTR-F1 & Dosage & Cosine & 31.2 / 16.5 & 31.1 / 16.5 & \textbf{31.7} / \textbf{20.1} & 31.3 / 17.8 & 29.2 / 15.6 & 29.0 / 16.1 & 29.7 / 15.6 & 23.5 / 12.9 \\
    % \midrule
    % --- DR
    DR & Syndrome & Accuracy & 86.7 / 39.8 & 85.6 / 35.5 & \textbf{86.9} / \textbf{41.5} & 85.4 / 40.8 & 86.0 / 39.3 & 83.6 / 30.5 & 84.9 / 32.8 & 77.7 / 34.8 \\
    DR & Treatment & Accuracy & 78.7 / 22.8 & 77.6 / 20.3 & 78.1 / 22.0 & 79.0 / 20.8 & \textbf{80.0} / 22.3 & 76.8 / 15.3 & 77.6 / 17.8 & 75.3 / \textbf{31.8} \\
    DR & Prescription & Accuracy & 89.1 / 51.2 & 90.0 / 55.3 & \textbf{90.6} / \textbf{56.8} & 88.7 / 49.3 & 87.5 / 45.3 & 87.9 / 44.0 & 88.4 / 47.0 & 78.5 / 37.3 \\
    % \midrule
    \textbf{Average} &  &  & \textbf{51.1} / \textbf{31.9} & 50.9 / 31.2 & 48.1 / 28.0 & 50.6 / 30.8 & 48.4 / 28.8 & 47.1 / 26.2 & 44.9 / 22.6 & 40.7 / 23.0 \\
    \bottomrule
    \end{tabular}
  }
  \par\smallskip
  \begin{minipage}{\textwidth}
    \footnotesize\raggedright
    Abbreviations: TLE (TCM Licensing Examination), FTK (Fundamental TCM Knowledge), CPMK (Chinese Patent Medicine Knowledge), IE (Information Extraction), DTR (Diagnostic \& Therapeutic Reasoning), DR (Decision Recognition).
  \end{minipage}
\end{table*}

% 小模型

\begin{table*}[t]
  \centering
  \scriptsize
  \setlength{\tabcolsep}{3pt}
  \caption{Small-scale LLM results covering all subtasks. Each entry reports full-set / Hard-subset performance (left/right). Bold values denote the highest score for each subtask–metric. \textit{Average} refers to the mean of all metrics, excluding MAE.}
  \label{tab:Small-LLM-performance}
  \resizebox{\textwidth}{!}{
  \begin{tabular}{lllccccccc}
    \toprule
    \textbf{Dom.} & \textbf{Subtask} & \textbf{Metric} & \textbf{Qwen3-32B} & \textbf{Qwen3-30B} & \textbf{Qwen3-14B} & \textbf{Qwen3-8B} & \textbf{Qwen3-4B} & \textbf{GPT-5-Mini} & \textbf{GPT-OSS-20B} \\
    \midrule
    % --- TLE
    TLE & Comprehensive & Accuracy & \textbf{91.2} / \textbf{62.3} & 88.8 / 55.3 & 87.6 / 50.0 & 84.6 / 40.5 & 76.6 / 30.3 & 76.3 / 24.5 & 48.3 / 13.5 \\
    % \midrule
    % --- FTK
    FTK & Single-choice & Accuracy & \textbf{86.1} / \textbf{44.3} & 82.6 / 32.0 & 83.0 / 34.5 & 80.8 / 33.8 & 76.2 / 28.5 & 77.0 / 28.7 & 65.4 / 29.3 \\
    FTK & Multiple-choice & Accuracy & \textbf{56.3} / \textbf{23.3} & 52.0 / 17.0 & 49.5 / 13.8 & 47.8 / 13.0 & 42.4 / 12.3 & 46.2 / 10.8 & 24.8 / 6.0 \\
    FTK & Multiple-choice & Precision & 90.3 / \textbf{77.5} & 89.8 / 75.2 & \textbf{90.6} / 75.2 & 88.4 / 71.8 & 86.1 / 69.0 & 86.1 / 67.7 & 76.1 / 59.5 \\
    FTK & Multiple-choice & Recall & \textbf{90.9} / \textbf{78.0} & 88.3 / 73.5 & 85.7 / 69.4 & 86.8 / 70.2 & 84.9 / 67.6 & 86.8 / 69.0 & 74.9 / 58.1 \\
    FTK & Multiple-choice & F1 & \textbf{89.1} / \textbf{74.8} & 87.4 / 71.0 & 86.3 / 68.7 & 85.8 / 67.6 & 83.5 / 64.8 & 84.8 / 65.3 & 72.9 / 55.0 \\
    FTK & Cloze & char-F1 & \textbf{56.1} / 17.5 & 53.9 / \textbf{17.9} & 54.1 / 14.3 & 51.1 / 12.3 & 49.0 / 15.9 & 49.2 / \textbf{17.9} & 35.9 / 10.6 \\
    % \midrule
    % --- CPMK
    CPMK & Single-choice & Accuracy & \textbf{66.9} / \textbf{16.3} & 58.4 / 5.5 & 58.9 / 9.8 & 56.6 / 9.0 & 51.3 / 9.3 & 51.8 / 9.5 & 37.6 / 11.8 \\
    CPMK & Multiple-choice & Accuracy & \textbf{35.5} / 6.5 & 34.7 / 5.8 & 34.4 / 5.3 & 31.4 / \textbf{7.2} & 28.7 / 4.5 & 28.9 / 4.0 & 18.5 / 3.8 \\
    CPMK & Multiple-choice & Precision & 81.9 / \textbf{62.1} & 81.8 / 60.6 & \textbf{82.5} / 61.5 & 81.3 / 61.6 & 79.3 / 59.3 & 78.2 / 54.9 & 72.6 / 51.1 \\
    CPMK & Multiple-choice & Recall & \textbf{87.8} / \textbf{70.0} & 85.4 / 65.1 & 84.9 / 64.6 & 84.2 / 66.3 & 84.7 / 67.7 & 82.7 / 57.9 & 75.4 / 55.4 \\
    CPMK & Multiple-choice & F1 & \textbf{83.2} / \textbf{63.8} & 82.0 / 60.7 & 82.0 / 60.4 & 80.9 / 61.6 & 80.0 / 60.7 & 78.7 / 54.3 & 71.6 / 50.8 \\
    CPMK & Cloze & char-F1 & \textbf{67.5} / 41.3 & 66.3 / 41.3 & 66.8 / 39.8 & 64.1 / 37.9 & 61.4 / 37.0 & 65.1 / \textbf{43.3} & 49.9 / 29.5 \\
    % \midrule
    % --- NER
    NER & Clinical EMR & Precision & 73.3 / 61.3 & 69.3 / 57.2 & \textbf{74.6} / \textbf{62.8} & 67.7 / 55.1 & 64.7 / 52.7 & 52.1 / 35.2 & 55.3 / 39.5 \\
    NER & Clinical EMR & Recall & \textbf{62.2} / 49.8 & 48.4 / 37.9 & 62.1 / \textbf{52.0} & 57.6 / 45.1 & 52.3 / 41.3 & 57.0 / 43.6 & 51.6 / 38.5 \\
    NER & Clinical EMR & F1 & 66.7 / 54.2 & 56.2 / 44.6 & \textbf{67.1} / \textbf{55.7} & 61.5 / 48.6 & 57.2 / 45.4 & 54.0 / 38.3 & 52.8 / 38.0 \\
    NER & Classical Texts & Precision & 57.9 / 38.1 & 59.7 / 38.9 & \textbf{59.8} / \textbf{39.6} & 58.1 / 35.4 & 54.5 / 35.0 & 48.9 / 23.9 & 50.3 / 26.6 \\
    NER & Classical Texts & Recall & 56.9 / 37.9 & 56.3 / 37.3 & 56.3 / \textbf{38.3} & 54.7 / 34.4 & 49.2 / 32.0 & \textbf{60.5} / 35.6 & 54.0 / 31.8 \\
    NER & Classical Texts & F1 & 56.6 / 36.9 & 57.2 / 37.0 & \textbf{57.3} / \textbf{38.0} & 55.6 / 34.0 & 51.0 / 32.4 & 53.4 / 28.0 & 51.3 / 28.1 \\
    % \midrule
    % --- DTR
    DTR & Syndrome & Precision & 7.7 / 0.6 & \textbf{8.4} / \textbf{1.4} & 7.6 / 1.2 & 7.5 / 1.0 & 6.9 / \textbf{1.4} & 3.4 / 0.3 & 3.3 / \textbf{1.4} \\
    DTR & Syndrome & Recall & 11.2 / 0.9 & \textbf{12.9} / 2.0 & 11.3 / 1.2 & 9.9 / 1.1 & 9.8 / 1.7 & 5.6 / 0.4 & 5.4 / \textbf{2.1} \\
    DTR & Syndrome & F1 & 8.8 / 0.7 & \textbf{9.8} / \textbf{1.6} & 8.8 / 1.2 & 8.2 / 1.0 & 7.8 / 1.5 & 4.1 / 0.3 & 3.9 / \textbf{1.6} \\
    DTR & Treatment & Precision & 11.5 / \textbf{1.3} & 13.0 / 0.8 & 11.0 / 0.8 & 12.0 / 0.8 & 11.6 / \textbf{1.3} & \textbf{13.4} / \textbf{1.3} & 6.5 / 0.8 \\
    DTR & Treatment & Recall & 14.4 / 1.5 & 15.4 / 1.1 & 14.3 / 1.1 & 14.7 / 1.0 & 14.4 / \textbf{1.7} & \textbf{15.7} / 1.5 & 7.8 / 0.9 \\
    DTR & Treatment & F1 & 12.3 / 1.3 & \textbf{13.6} / 0.9 & 12.0 / 0.9 & 12.8 / 0.9 & 12.4 / \textbf{1.5} & 12.9 / 1.2 & 6.7 / 0.8 \\
    DTR & Prescription & Precision & 31.0 / 17.0 & \textbf{32.4} / \textbf{18.1} & 31.1 / 17.2 & 31.7 / 17.1 & 30.4 / 16.7 & 27.6 / 16.1 & 24.8 / 15.7 \\
    DTR & Prescription & Recall & 29.3 / 16.9 & 26.0 / 14.9 & 29.6 / 17.2 & 26.4 / 14.7 & 25.0 / 14.0 & \textbf{30.8} / \textbf{19.0} & 21.2 / 14.0 \\
    DTR & Prescription & F1 & 29.5 / 16.4 & 28.3 / 15.8 & \textbf{29.8} / 16.7 & 28.2 / 15.2 & 26.8 / 14.7 & 28.5 / \textbf{16.9} & 22.2 / 14.2 \\
    DTR & Dosage & MAE & 4.4 / 4.7 & 4.4 / \textbf{4.0} & 4.2 / 4.2 & 4.3 / 4.1 & \textbf{4.1} / 4.1 & 5.3 / 5.3 & 4.5 / 4.3 \\
    DTR & Dosage & Cosine & \textbf{29.0} / 15.5 & 28.1 / 15.3 & 28.6 / 15.7 & 27.2 / 14.2 & 26.3 / 14.5 & 27.9 / \textbf{16.0} & 20.0 / 12.4 \\
    % \midrule
    % --- DTR-F1
    DTR-F1 & Syndrome & Precision & 16.5 / 7.8 & 17.3 / \textbf{9.2} & 16.5 / 8.3 & \textbf{17.4} / 8.2 & 15.5 / 8.1 & 10.1 / 4.7 & 7.7 / 3.4 \\
    DTR-F1 & Syndrome & Recall & 23.8 / 10.5 & \textbf{26.5} / \textbf{13.0} & 24.1 / 11.3 & 22.5 / 10.2 & 21.9 / 10.7 & 16.0 / 6.4 & 11.9 / 4.7 \\
    DTR-F1 & Syndrome & F1 & 18.7 / 8.7 & \textbf{20.1} / \textbf{10.4} & 18.8 / 9.2 & 18.8 / 8.8 & 17.5 / 8.8 & 11.9 / 5.3 & 8.8 / 3.8 \\
    DTR-F1 & Treatment & Precision & 18.0 / 6.7 & \textbf{20.1} / \textbf{7.7} & 16.8 / 6.2 & 18.8 / 6.1 & 17.4 / 6.2 & 17.2 / 4.0 & 10.3 / 3.8 \\
    DTR-F1 & Treatment & Recall & 23.0 / 9.0 & \textbf{23.9} / \textbf{10.2} & 22.4 / 8.8 & 23.4 / 8.5 & 21.8 / 8.6 & 22.6 / 7.1 & 13.4 / 5.4 \\
    DTR-F1 & Treatment & F1 & 19.5 / 7.5 & \textbf{21.1} / \textbf{8.6} & 18.6 / 7.1 & 20.2 / 6.9 & 18.8 / 7.1 & 17.4 / 4.8 & 11.1 / 4.3 \\
    DTR-F1 & Prescription & Precision & 30.9 / 17.0 & \textbf{32.4} / \textbf{18.1} & 30.9 / 17.1 & 31.7 / 17.2 & 30.4 / 16.8 & 27.7 / 16.3 & 25.4 / 15.7 \\
    DTR-F1 & Prescription & Recall & 29.6 / 17.0 & 26.0 / 14.9 & 29.9 / 17.4 & 26.5 / 14.8 & 25.1 / 14.1 & \textbf{31.1} / \textbf{19.2} & 21.9 / 14.2 \\
    DTR-F1 & Prescription & F1 & 29.6 / 16.5 & 28.3 / 15.8 & \textbf{29.8} / 16.8 & 28.3 / 15.3 & 26.9 / 14.8 & 28.7 / \textbf{17.1} & 22.9 / 14.4 \\
    DTR-F1 & Dosage & MAE & 4.4 / 4.7 & 4.4 / \textbf{4.0} & 4.2 / 4.3 & 4.3 / 4.1 & \textbf{4.1} / 4.1 & 5.3 / 5.4 & 4.6 / 4.4 \\
    DTR-F1 & Dosage & Cosine & \textbf{29.2} / 15.6 & 28.2 / 15.3 & 28.7 / 15.8 & 27.4 / 14.5 & 26.3 / 14.6 & 28.1 / \textbf{16.2} & 20.8 / 12.5 \\
    % \midrule
    % --- DR
    DR & Syndrome & Accuracy & \textbf{86.0} / \textbf{39.3} & 84.1 / 29.5 & 84.2 / 34.5 & 82.8 / 29.0 & 77.3 / 25.3 & 80.7 / 26.0 & 70.3 / 30.3 \\
    DR & Treatment & Accuracy & 80.0 / 22.3 & \textbf{80.4} / 23.8 & 78.4 / 19.0 & 78.6 / 20.3 & 74.8 / 24.0 & 78.1 / 26.0 & 68.2 / \textbf{33.5} \\
    DR & Prescription & Accuracy & \textbf{87.5} / \textbf{45.3} & 86.7 / 40.5 & 87.4 / 43.3 & 84.9 / 35.3 & 82.4 / 36.0 & 84.4 / 39.5 & 64.5 / 25.5 \\
    % \midrule
    \textbf{Average} &  &  & \textbf{48.4} / \textbf{28.8} & 47.2 / 26.7 & 47.5 / 27.2 & 46.2 / 25.4 & 43.8 / 24.5 & 43.8 / 23.3 & 36.1 / 20.8 \\
    \bottomrule
    \end{tabular}
  }
  \par\smallskip
  \begin{minipage}{\textwidth}
    \footnotesize\raggedright
    Abbreviations: TLE (TCM Licensing Examination), FTK (Fundamental TCM Knowledge), CPMK (Chinese Patent Medicine Knowledge), IE (Information Extraction), DTR (Diagnostic \& Therapeutic Reasoning), DR (Decision Recognition).
  \end{minipage}
\end{table*}

\subsection{Overall Performance Across Domains}

Licensing–style questions attain the highest accuracies on both the full sets and the hard subsets. In FTK and CPMK, single-choice accuracy is lower than per-option F1 on multi-choice, and cloze scores are lower than multi-choice; degradations on the hard subsets are larger for single-choice and cloze than for multi-choice. For information extraction, EMR scores exceed classical-text scores on both settings, and the gap persists on the hard subsets. In diagnostic–therapeutic reasoning, strict exact matching (DTR) yields lower scores than the synonym-tolerant variant (DTR-F1) across syndrome, treatment, and prescription, while reformulating the same problems as decision recognition (DR) produces the highest accuracies among these formats. For dosage estimation, cosine similarity varies more across systems, whereas MAE remains within a comparatively narrow range on both the full and hard sets.

\subsection{Overall Performance Across Models}

On the full sets, overall averages cluster into three bands: a leading group centered near 51 (DeepSeek-R1, DeepSeek-V3.1, Qwen3-235B), a middle band in the high 40s (Qwen3-32B, GPT-5, Baichuan-M2-32B), and a lower band in the mid to low 40s (Qwen3-Next-80B, GPT-OSS-120B). On the hard subsets, the corresponding means concentrate near 31, with absolute declines averaging about 20 points across models. DeepSeek-V3.1 achieves the highest licensing-exam accuracy and the strongest CPMK cloze on the full set; DeepSeek-R1 attains the highest multiple-choice F1 in FTK and CPMK; Qwen3-235B leads FTK cloze; GPT-5 attains the top EMR extraction F1, the highest DR prescription accuracy, and the best dosage cosine.

\subsection{Hard Subset Evaluation}

All systems show pronounced performance declines on the hard subsets across domains, with the magnitude varying by task type and model. Among knowledge-oriented tasks, FTK multi-choice accuracy commonly declines by roughly 20–35 absolute points, and FTK cloze character-F1 drops by about 30–40 points. In CPMK, multi-choice F1 decreases by about 15–20 points, while CPMK cloze exhibits a smaller but still substantial contraction. 
Information extraction shows moderate-to-large declines: EMR F1 typically falls by 10–17 points, whereas classical-text F1 drops by about 15–30 points. 
In the diagnostic–therapeutic suite, strict DTR prescription F1 contracts from the low 30s to the high teens; dosage cosine similarity decreases by roughly 10–12 points. The decision-recognition reformulation (DR) remains the most accurate format on the full sets but also suffers the largest absolute drops in some subtasks, with syndrome accuracy falling from the high 80s to the low 40s, treatment often declining from around 80 to the low 20s, and prescription from about 90 to the mid-50s.
Model-wise, high-capacity systems generally retain stronger hard-subset performance, yet all experience notable erosion.
GPT-5 exhibits asymmetric robustness, leading EMR extraction but dropping sharply on classical-text extraction. Mid-tier and smaller models show steeper relative declines in several knowledge and extraction subtasks; for instance, GPT-OSS-120B and GPT-OSS-20B register large losses on TLE and multi-choice QA. Taken together, the hard-subset results consistently stress models across knowledge recall, span extraction, multi-label clinical reasoning, and dosage proportionality, providing a stringent lens on robustness beyond full-set performance.

\section{Discussion}

LingLanMedBench reveals a consistent gap between TCM knowledge recall and clinically faithful reasoning. Across models, licensing-style and foundational knowledge tasks show comparatively strong performance, whereas hard subsets, classical-text extraction, open-ended syndrome-treatment reasoning, prescription composition, and dosage proportionality remain substantially more difficult. This pattern suggests that current LLMs can retrieve and recognize TCM knowledge more reliably than they can compose it under realistic clinical constraints.

Compared with TCMD~\cite{yu2024tcmd}, TCMBench~\cite{yue2024tcmbench}, MTCMB~\cite{kong2025mtcmb}, and TCMEval-SDT~\cite{wang2025tcmeval}, LingLanMedBench extends evaluation from examination-style knowledge and isolated generation tasks to a broader suite covering foundational knowledge, Chinese patent medicine, clinical information extraction, diagnostic and therapeutic reasoning, and decision recognition. In contrast to general biomedical benchmarks, LingLanMedBench targets TCM-specific ontology, terminology, syndrome-based reasoning, prescription composition, and dosage proportionality. This positioning is summarized in Table~\ref{tab:tcm_benchmarks}.

\begin{table*}[!t]
    \centering
    \small
    \setlength{\tabcolsep}{4pt}
    \renewcommand{\arraystretch}{1.2}
    \caption{Comparison of major TCM benchmarks by scale, task types, sources, and metrics.}
    \label{tab:tcm_benchmarks}
    \resizebox{\textwidth}{!}{%
    \begin{tabular}{@{}p{2.7cm}p{2.5cm}p{3.5cm}p{3.8cm}p{4.2cm}@{}}
    \toprule
    \textbf{Benchmark} & \textbf{Scale} & \textbf{Question Type} & \textbf{Sources} & \textbf{Metrics} \\ 
    \midrule
    \textbf{TCMD} \cite{yu2024tcmd} & 3,451 & MC & Licensing examinations & Accuracy \\
    \textbf{TCMBench} \cite{yue2024tcmbench} & 5,473 & MC & Licensing examinations & TCMScore, ROUGE, BERTScore \\
    \textbf{MTCMB} \cite{kong2025mtcmb} & 7,100 (6,000 exam samples) & MC / Cloze / Open QA / Prescription / Diagnosis / IE & Licensing examinations / Classical texts / Clinical records / Guidelines & Accuracy, ROUGE, BLEU, BERTScore \\
    \textbf{TCMEval-SDT} \cite{wang2025tcmeval} & 300 cases & IE / Pathogenesis / Syndrome / Summary & Internet / Classical texts / Clinical records & Stage-wise qualitative \\
    \textbf{LingLanMedBench (ours)} & 25,620 ( \textasciitilde 2,000 per subtask) & SC / MC / Cloze / IE / DTR / DR & TLE / FTK / CPMK / EMR / Classical texts / Master-physician casebooks & Accuracy, Precision, Recall, Char-F1, MAE, Cosine similarity \\
    \bottomrule
    \end{tabular}
}
    \par\smallskip
    \begin{minipage}{\textwidth}
      \footnotesize\raggedright
      Abbreviations: SC (single-choice), MC (multiple-choice), Cloze (fill-in-the-blank), IE (information extraction), DTR (diagnostic–therapeutic reasoning), DR (decision recognition), TLE (TCM Licensing Examination), FTK (Fundamental TCM Knowledge), CPMK(Chinese Patent Medicine Knowledge), EMR(electronic medical records), Char-F1(character-level F1), MAE(mean absolute error).
    \end{minipage}
\end{table*}

\subsection{Impact of Model Scale and Architecture}

The results further indicate that model scale improves performance unevenly across task formats.
On knowledge-oriented tasks (TLE, FTK, CPMK), performance increases clearly with scale: larger models better maintain stem–option alignment and terminology recall, especially in multi-choice settings, yielding higher and more stable option-level precision/recall/F1.
This pattern suggests that “parameterized memory” and option calibration benefit from greater capacity. 

In contrast, IE does not improve monotonically with size: smaller models sometimes exceed larger ones in macro-F1 on both EMR and classical-text corpora, indicating that extraction accuracy is sensitive to modeling behavior rather than raw capacity alone. Under the IE protocol (exact match on entity type, surface string with no synonym merging or normalization), larger models frequently exhibit benevolent normalization (standardizing terms, adjusting punctuation, consolidating near-duplicates) and confident inference (widening span boundaries or adding implicitly suggested entities). These behaviors reduce exact-match rates and can introduce type or polarity (positive/negative) mismatches, lowering precision and recall. Smaller models tend to copy spans verbatim and keep tighter boundaries, aligning more closely with the scoring criterion and yielding steadier macro-F1 despite their lower overall capacity.

Along the syndrome, treatment, prescription, dosage chain (DTR task), a common bottleneck persists across scales: enforcing global consistency over multi-label sets and exercising compositional control over prescriptions. Gains from scaling are modest; when reframed as single-choice DR, the search space contracts and size advantages re-emerge, yet all models still drop markedly on the hard subset, indicating robustness and transfer have not kept pace. Architecturally, dense models show better overall balance on knowledge recall and extraction, while mixture-of-experts (MoE) models tend to gain marginal advantages on discrete category decisions (e.g., syndrome, treatment method), likely from expert routing; these gains do not extend to strongly compositional or continuous-space tasks such as prescription composition and dosage ratios.

Overall, scale chiefly improves memory-driven, output-constrained formats, whereas architectural effects are task-dependent—benefiting categorical judgment but remaining insensitive to compositional structure. Continued progress on extraction and compositional reasoning will require tighter terminology and boundary standards, synonym-aware matching, and explicit mechanisms for chain consistency and prescription control.

\subsection{Knowledge-Oriented Tasks}

Knowledge-oriented tasks show that strong full-set accuracy does not necessarily indicate robust TCM competence.

On the full sets, top systems approach ceiling accuracy on the TLE. A plausible explanation is that pretraining corpora may contain historical exam items or close paraphrases, which enhances recall of taught curricula and familiar test formats. Notably, this is the only LingLanMedBench component that could plausibly overlap with web-accessible material; all other datasets are sourced from non-web resources. Even so, the hard subset still depresses accuracy and exposes sensitivity to carefully engineered distractors and cross-chapter integrations.
For multiple choice, a consistent gap appears between instance-level accuracy and per-option F1. Models often identify many correct options but miscalibrate the predicted subset size, leading to over-selection or under-selection. 
This highlights option calibration, rather than knowledge coverage alone, as a key challenge that LingLanMedBench makes explicit by reporting both metrics.
Cloze questions are evaluated with character-level F1, which tolerates minor lexical and formatting variation relative to exact match, reducing spurious penalties from near-synonyms while better reflecting underlying knowledge. The metric remains deterministic and reproducible, supporting fair cross-model comparison.

Across these knowledge-oriented tasks, Chinese-centric model families (e.g., DeepSeek and Qwen) consistently occupy the top tier. This pattern is consistent with several language-domain factors: higher coverage of TCM terminology (materia medica, formula names, syndrome lexicon) in their pretraining data; tokenizers optimized for Chinese scripts that reduce sequence fragmentation; and instruction tuning that emphasizes Chinese exam formats and short-form factual recall. 
The advantage is most pronounced on TLE, FTK, and CPMK, which are dominated by recall and recognition, while it narrows in settings that demand span-faithful extraction or compositional generation.

\subsection{Information Extraction Tasks}

The information extraction task comprises two subtasks: NER on de-identified EMRs and on classical TCM texts in Literary Chinese.

\subsubsection{Genre Effects and Scoring Considerations}

Information extraction exposes a second evaluation axis: genre robustness. Modern EMRs are relatively tractable because they use semi-structured contemporary clinical prose, whereas classical TCM texts introduce archaic vocabulary, elliptical syntax, and long-distance dependencies that make span delimitation and entity typing more fragile. The instance-level multiset scoring protocol further makes over-extraction, under-extraction, and duplicate handling visible rather than hiding them through set-level aggregation.

\subsubsection{GPT: Strong on EMR, Weak on Classical Texts}

Across models, GPT-5 yields the best F1 on EMR, reflecting effective adaptation to contemporary clinical Chinese and stable span decisions under distribution shift. On classical texts, however, GPT-5 falls behind models such as DeepSeek-V3.1, which better handle archaic vocabulary and compact syntax. The Qwen3 family remains competitive across both genres and typically shows higher precision than recall on EMR, consistent with a cautious extraction style. Models with limited Chinese clinical and classical coverage, exemplified by GPT-OSS-120B, underperform on both corpora. Models whose pretraining is predominantly Chinese, such as DeepSeek and Qwen, tend to generalize better to classical Chinese, likely due to broader exposure to historical corpora, tokenization optimized for Chinese characters, and instruction tuning on Chinese sources.

\subsubsection{Hard-Subset Effects and Methodological Directions}

Degradation on the hard subsets is systematic and more pronounced for classical texts than for EMR, reflecting rarer terminology, compositional symptom phrases, and atypical surface forms that complicate span boundaries and type assignment. Improving robustness will likely require genre-aware tokenization and prompting for classical Chinese, span decision mechanisms that incorporate broader context without inflating boundaries, and post hoc consistency checks that reconcile entity inventories within long documents while balancing precision and recall.

\subsection{Diagnostic and Therapeutic Reasoning Tasks}

The diagnostic and therapeutic suite contains four interdependent subtasks in a free-form setting (DTR) and a recognition setting (DR) that reformulates the first three as single-choice questions.

\subsubsection{Free-form Composition Remains Challenging}
Across models, strict DTR scores are low in absolute terms and drop further on the hard subset, which indicates that open-set composition remains challenging. Under exact matching, prescription F1 peaks in the low thirties on the full set and falls to the high teens on the hard subset. Introducing the synonym-tolerant protocol yields modest but consistent gains, suggesting that a nontrivial share of errors arises from surface variation rather than conceptual mismatch. Error profiles show a stable precision–recall tradeoff: larger-capacity systems tend to retrieve more relevant labels, which improves recall but also introduces additional, partially plausible items that depress precision; smaller systems are more conservative, which stabilizes precision but limits coverage of multi-component prescriptions. Typical failure modes include under-selection of secondary herbs, confusion between closely related treatment principles, and incomplete alignment among syndrome, treatment, and prescription when long-range symptom relations are required.

Dosage estimation exhibits a different pattern. Cosine similarity, which emphasizes proportional agreement among dosed herbs, ranges in the low thirties on the full set and declines by roughly ten points on the hard subset. Mean absolute error remains around four to five grams per herb on both splits, with smaller values indicating better absolute calibration. The two metrics emphasize complementary behaviors. Cosine similarity is informative when overlap between predicted and reference herbs is limited, since it rewards correct dose ratios even with partial matches. MAE becomes more discriminative when most herbs are matched, since it measures absolute deviation. Observed errors concentrate on global scaling (for example, near-correct proportions but uniformly larger or smaller doses), unit normalization inconsistencies across sources, and occasional outlier herbs with atypical gram ranges that dominate the MAE.

\subsubsection{Recognition Reformulation Improves Reliability}

Recasting the same clinical problems as decision recognition substantially raises accuracy. With a constrained option set constructed from nearest-neighbor cases, models align more closely with expert labels for syndrome, treatment, and prescription. Hard-subset accuracy still drops, but the gap relative to DTR indicates that a significant portion of difficulty in free-form reasoning stems from the size of the hypothesis space and from calibration over multi-label outputs, rather than from a lack of stored knowledge alone. This pattern suggests practical routes to improvement, including candidate generation and re-ranking pipelines, structure-aware decoding that enforces compatibility among syndrome, treatment, and prescription, and training objectives that penalize over-selection and reward proportionally correct dosage vectors.

\subsection{Terminology Normalization for Reliable TCM Evaluation}
Existing studies show that general-purpose LLMs can align closely with expert judgments on TCM-specific tasks \cite{liu2025evaluating}; however, this evidence is narrow and depends on single-case, manual adjudication.
A central impediment is terminology: pervasive synonymy, variant canonical forms, and heterogeneous granularity in syndrome labels, treatment principles, herb names, and processing states complicate both human and automated scoring. 
The field currently lacks a synonym and normalization resource broad enough to support reliable evaluation across these tasks.
In LingLanMedBench, we partially mitigate lexical variability by using character-level F1 for surface-form tolerance and by reformulating open-ended outputs as constrained decision recognition, which improves comparability without overfitting to any single phrasing.

\section{Conclusion}
LingLanMedBench provides a large-scale, expert-curated benchmark for evaluating LLMs across TCM knowledge, information extraction, and diagnostic--therapeutic decision-making under a unified protocol. Evaluation of fourteen contemporary LLMs reveals a persistent gap between strong performance on knowledge-oriented tasks and substantially weaker performance on open-ended clinical reasoning, particularly under difficult settings. Reformulating clinical decisions as recognition tasks markedly improves performance, indicating that output-space complexity and calibration contribute substantially to current limitations. By standardizing heterogeneous tasks and evaluation metrics, LingLanMedBench provides a reproducible foundation for characterizing model capabilities and tracking progress toward more clinically faithful TCM reasoning.
LingLanMedBench is currently text-only and therefore does not capture multimodal cues central to TCM practice, including tongue, facial, and pulse information. Future extensions should incorporate these modalities with standardized expert annotations to enable reproducible evaluation of multimodal TCM reasoning.

\section{Acknowledgements}

We thank Shujie Song, Mingzhong Xiao, Naixuan Zhao, Yu Fu, Benman Yu, Yuanyuan Chen, Zhizhou Meng, Pengcheng Yang, Songxue Zhao, Lijuan Pei, Yunhui Hu and Yang Gu for their contributions to data curation and validation. We thank Zeyu Liu, Jiayuan Duan and Wenmao Yin for software development and technical support. We thank Qiang Zhu for scientific consultation and methodological advice.

This work is partially supported by the National Key Research and Development Program (No. 2023YFC3502604), the Natural Science Foundation of Beijing (No. L232033), the National Natural Science Foundation of China (Nos. 82374302, 82274352, 82505376 and U23B2062), the Fundamental Research Funds for the Central Universities (No. 2025JBZX064).

\section{Author Contributions}

RH: Investigation, Writing – original draft.
YW, ZS, HZ, XL, DJ, ZG, YM, HY, FD, DY, KD and RZ: Data curation, Validation.
JX, KC: Methodology.
JY and JL: Conceptualization, Methodology, Supervision.
WW: Resources, Writing – review \& editing.
BL and XZ: Conceptualization, Methodology, Supervision, Writing – review \& editing.
All authors reviewed and approved the final manuscript.

\section{Data availability and compliance statement}

The LingLanMedBench benchmark was constructed from traditional Chinese medicine licensing examination questions, an in-house knowledge
graph integrating classical texts, textbooks, biomedical literature, SymMap and TCMID, the Pharmacopoeia of the People's Republic of China, package inserts of Chinese patent medicines, classical medical literature, de-identified clinical medical records, and case records of traditional Chinese medicine physicians. The data sources and curation procedures are described in the Materials and methods section.

The benchmark input files and evaluation scripts are publicly available in the LingLanMedBench repository at \url{https://github.com/TCMAI-BJTU/LingLan}. The reference annotations required to run the evaluation and reproduce the reported results are not yet publicly available. The complete evaluation datasets, including these annotations, will be released in the same repository upon acceptance of this manuscript.

The authors affirm that all data used in this study were obtained and used in accordance with applicable legislation, regulatory requirements, ethical standards, and the terms governing the original data sources. The authors take responsibility for the lawful acquisition and use of these data.

\section{Compliance with ethics guidelines}

\subsection{Conflicts of interest}

Rui Hua, Yu Wei, Zixin Shu, Kai Chang, Dengying Yan, Jianan Xia, Hui Zhu, Xiaodong Li, Dongmei Jia, Zhuye Gao, Yanyan Meng, Haibin Yu, Fei Dong, Kan Ding, Runshun Zhang, Jian Yu, Jiansheng Li, Baoyan Liu, Wenjia Wang and Xuezhong Zhou declare that they have no conflict of interest.

\subsection{Ethics approval and consent to participate}

Not applicable.

\subsection{Consent for publication}

Not applicable.

%Bibliography
\bibliographystyle{unsrt}
\bibliography{references}

% \documentclass{article}

% \usepackage[utf8]{inputenc}
% \usepackage[T1]{fontenc}
% \usepackage{PRIMEarxiv}
% \usepackage{graphicx,booktabs,amsmath,microtype}
% \usepackage{lineno}
% \usepackage[hidelinks]{hyperref}

% % Clear the template header and retain page numbers.
% \fancyhf{}
% \cfoot{\thepage}

% \begin{document}

% \clearpage
% \nolinenumbers

\clearpage
\nolinenumbers

% Clear the template header and retain continuous page numbers.
\fancyhf{}
\cfoot{\thepage}

\section*{Supplementary material}
% Number supplementary sections as S1, S2, etc.
\setcounter{subsection}{0}
\renewcommand{\thesubsection}{S\arabic{subsection}}
\renewcommand{\theHsubsection}{supp.\arabic{subsection}}

% Number supplementary figures as S1, S2, etc.
\setcounter{figure}{0}
\renewcommand{\thefigure}{S\arabic{figure}}
\renewcommand{\theHfigure}{supp.\arabic{figure}}

% Number supplementary tables as S1, S2, etc.
\setcounter{table}{0}
\renewcommand{\thetable}{S\arabic{table}}
\renewcommand{\theHtable}{supp.\arabic{table}}

\subsection{Evaluation on TCM LLMs}

We evaluate six models under a unified zero-shot protocol with identical decoding: a general-purpose baseline (Qwen3-14B); a Chinese medical model (Baichuan2-13B-Chat); two TCM-specific systems fine-tuned from Baichuan2-13B-Chat (TCMChat, Lingdan-13B-TCPM); and two TCM-specific systems fine-tuned from Qwen2.5 (BianCang-7B, BianCang-14B), with BianCang-7B supporting an 8,192-token maximum context length. Qwen3-14B attains the highest overall averages on both full and hard splits (Table~\ref{tab:TCM-LLM-performance}) and leads NER on EMR and classical corpora. BianCang-7B and BianCang-14B are competitive on licensing-style assessment and decision recognition, especially on hard subsets, but trail Qwen3-14B on structured extraction and strict diagnostic–therapeutic composition. TCMChat and Lingdan-13B-TCPM show limited capability beyond single-choice QA, consistent with format-centric fine-tuning and the 4,096-token context limit inherited from Baichuan2-13B-Chat. Current TCM-specific models do not provide a dedicated reasoning mode, which further contributes to the substantial performance gap relative to the Qwen3 family. In dosage evaluation, near-zero MAE for some TCM models reflects failure to align herb names under inclusion-based matching and should not be interpreted as perfect accuracy; matching rules are detailed in Section~\ref{sec:supp_metrics}.

\begin{table*}[b]
  \centering
  \scriptsize
  \setlength{\tabcolsep}{3pt}
  \caption{TCM LLM results covering all domains and subtasks. Each entry reports full-set / Hard-subset performance (left/right). Bold values denote the highest score for each subtask–metric.}
  \label{tab:TCM-LLM-performance}
  % \resizebox{\textwidth}{!}{
  \begin{tabular}{lllcccccc}
    \toprule
    \textbf{Dom.} & \textbf{Subtask} & \textbf{Metric} & \textbf{Qwen3-14B} & \textbf{Baichuan2-13B-Chat} & \textbf{Lingdan-13B-TCPM} & \textbf{TCMChat} & \textbf{BianCang-7B} & \textbf{BianCang-14B} \\
    \midrule
    % --- TLE
    TLE & Comprehensive & Accuracy & 87.6 / 50.0 & 50.1 / 21.8 & 44.3 / 20.8 & 31.6 / 19.5 & 88.9 / 70.0 & \textbf{91.2} / \textbf{73.5} \\
    \midrule
    % --- FTK
    FTK & Single-choice & Accuracy & \textbf{83.0} / 34.5 & 60.6 / 28.5 & 48.4 / 26.3 & 11.1 / 7.5 & 78.2 / 43.3 & 82.6 / \textbf{49.3} \\
    FTK & Multiple-choice & Accuracy & \textbf{49.5} / 13.8 & 23.4 / 12.8 & 14.3 / 9.5 & 0.6 / 0.5 & 36.9 / \textbf{19.5} & 23.7 / 13.0 \\
    FTK & Multiple-choice & Precision & \textbf{90.6} / \textbf{75.2} & 71.8 / 62.0 & 53.6 / 48.8 & 58.9 / 48.2 & 78.9 / 70.1 & 65.2 / 59.8 \\
    FTK & Multiple-choice & Recall & \textbf{85.7} / 69.4 & 84.4 / \textbf{79.8} & 56.5 / 56.4 & 36.3 / 29.1 & 78.4 / 72.7 & 53.1 / 48.7 \\
    FTK & Multiple-choice & F1 & \textbf{86.3} / \textbf{68.7} & 75.9 / 67.7 & 52.4 / 49.0 & 39.9 / 32.4 & 76.2 / 68.0 & 55.8 / 50.3 \\
    FTK & Cloze & char-F1 & \textbf{54.1} / 14.3 & 33.2 / 13.3 & 8.5 / 4.5 & 13.9 / 6.5 & 35.4 / \textbf{15.0} & 38.7 / 14.6 \\
    \midrule
    % --- CPMK
    CPMK & Single-choice & Accuracy & 58.9 / 9.8 & 41.6 / 24.8 & 28.3 / 16.3 & 19.2 / 18.5 & 58.4 / 31.8 & \textbf{62.5} / \textbf{35.5} \\
    CPMK & Multiple-choice & Accuracy & \textbf{34.4} / 5.3 & 16.8 / 5.3 & 9.7 / 3.5 & 0.8 / 0.8 & 24.7 / \textbf{11.3} & 14.1 / 4.5 \\
    CPMK & Multiple-choice & Precision & \textbf{82.5} / 61.5 & 71.5 / 60.0 & 51.3 / 48.1 & 55.1 / 46.4 & 75.5 / \textbf{64.1} & 39.0 / 31.1 \\
    CPMK & Multiple-choice & Recall & \textbf{84.9} / 64.6 & 83.5 / 76.5 & 57.9 / 59.2 & 28.1 / 25.2 & 83.0 / \textbf{77.8} & 36.6 / 28.3 \\
    CPMK & Multiple-choice & F1 & \textbf{82.0} / 60.4 & 75.2 / 65.4 & 52.3 / 51.2 & 33.8 / 29.7 & 77.1 / \textbf{68.3} & 36.3 / 27.8 \\
    CPMK & Cloze & char-F1 & \textbf{66.8} / 39.8 & 37.2 / 26.4 & 11.7 / 9.3 & 13.8 / 10.6 & 54.4 / 37.5 & 58.4 / \textbf{42.2} \\
    \midrule
    % --- NER
    NER & Clinical EMR & Precision & \textbf{74.6} / \textbf{62.8} & 27.0 / 25.0 & 1.2 / 1.1 & 0.0 / 0.1 & 25.5 / 21.7 & 38.1 / 33.2 \\
    NER & Clinical EMR & Recall & \textbf{62.1} / \textbf{52.0} & 25.0 / 22.6 & 0.8 / 0.9 & 0.0 / 0.0 & 15.0 / 11.5 & 26.1 / 21.5 \\
    NER & Clinical EMR & F1 & \textbf{67.1} / \textbf{55.7} & 25.1 / 22.7 & 0.8 / 0.8 & 0.0 / 0.0 & 16.8 / 12.9 & 29.5 / 24.8 \\
    NER & Classical Texts & Precision & \textbf{59.8} / \textbf{39.6} & 32.3 / 21.5 & 5.6 / 2.8 & 1.5 / 1.8 & 46.1 / 25.2 & 55.3 / 39.0 \\
    NER & Classical Texts & Recall & \textbf{56.3} / \textbf{38.3} & 27.8 / 18.1 & 2.9 / 1.5 & 0.2 / 0.2 & 44.9 / 26.9 & 46.0 / 32.4 \\
    NER & Classical Texts & F1 & \textbf{57.3} / \textbf{38.0} & 29.0 / 18.7 & 3.4 / 1.7 & 0.3 / 0.4 & 44.1 / 24.5 & 49.0 / 34.0 \\
    \midrule
    % --- DTR
    DTR & Syndrome & Precision & 7.6 / 1.2 & 4.0 / 1.5 & 1.0 / 0.0 & 0.7 / 0.2 & 5.8 / 1.5 & \textbf{9.4} / \textbf{1.7} \\
    DTR & Syndrome & Recall & \textbf{11.3} / 1.2 & 5.2 / \textbf{1.3} & 0.6 / 0.0 & 0.8 / 0.3 & 4.7 / 0.9 & 6.5 / 1.0 \\
    DTR & Syndrome & F1 & \textbf{8.8} / 1.2 & 4.3 / \textbf{1.3} & 0.7 / 0.0 & 0.7 / 0.3 & 4.9 / 1.1 & 7.3 / 1.2 \\
    DTR & Treatment & Precision & 11.0 / 0.8 & 7.9 / 1.5 & 1.3 / 0.1 & 3.6 / 0.8 & 9.0 / 1.3 & \textbf{12.4} / \textbf{1.8} \\
    DTR & Treatment & Recall & \textbf{14.3} / 1.1 & 7.6 / \textbf{1.4} & 0.9 / 0.1 & 2.6 / 0.5 & 6.3 / 1.0 & 7.8 / 1.0 \\
    DTR & Treatment & F1 & \textbf{12.0} / 0.9 & 7.4 / \textbf{1.3} & 1.0 / 0.1 & 2.8 / 0.6 & 7.1 / 1.1 & 9.1 / 1.2 \\
    DTR & Prescription & Precision & \textbf{31.1} / \textbf{17.2} & 15.8 / 9.5 & 0.3 / 0.2 & 0.0 / 0.0 & 15.0 / 9.8 & 8.8 / 5.4 \\
    DTR & Prescription & Recall & \textbf{29.6} / \textbf{17.2} & 12.7 / 8.1 & 0.2 / 0.2 & 0.0 / 0.0 & 15.4 / 10.1 & 8.3 / 5.6 \\
    DTR & Prescription & F1 & \textbf{29.8} / \textbf{16.7} & 13.6 / 8.3 & 0.2 / 0.2 & 0.0 / 0.0 & 14.4 / 9.1 & 8.2 / 5.1 \\
    DTR & Dosage & MAE & 4.2 / 4.2 & 2.3 / 2.2 & 0.1 / \textbf{0.0} & \textbf{0.0} / \textbf{0.0} & 2.5 / 2.0 & 1.4 / 1.3 \\
    DTR & Dosage & Cosine & \textbf{28.6} / \textbf{15.7} & 12.9 / 7.5 & 0.2 / 0.1 & 0.0 / 0.0 & 13.5 / 8.6 & 7.5 / 4.9 \\
    \midrule
    % --- DTR-F1
    DTR-F1 & Syndrome & Precision & 16.5 / 8.3 & 8.8 / 5.5 & 1.6 / 0.1 & 0.8 / 0.5 & 12.8 / 7.8 & \textbf{19.9} / \textbf{8.4} \\
    DTR-F1 & Syndrome & Recall & \textbf{24.1} / \textbf{11.3} & 10.9 / 6.2 & 1.1 / 0.1 & 0.9 / 0.6 & 10.0 / 5.2 & 13.7 / 4.5 \\
    DTR-F1 & Syndrome & F1 & \textbf{18.8} / \textbf{9.2} & 9.4 / 5.6 & 1.2 / 0.1 & 0.8 / 0.5 & 10.6 / 6.0 & 15.4 / 5.5 \\
    DTR-F1 & Treatment & Precision & 16.8 / 6.2 & 11.6 / 3.9 & 1.5 / 0.3 & 4.1 / 0.9 & 14.2 / 4.6 & \textbf{19.4} / \textbf{7.9} \\
    DTR-F1 & Treatment & Recall & \textbf{22.4} / \textbf{8.8} & 11.2 / 4.2 & 1.0 / 0.2 & 2.9 / 0.7 & 9.8 / 3.6 & 12.3 / 4.8 \\
    DTR-F1 & Treatment & F1 & \textbf{18.6} / \textbf{7.1} & 10.9 / 3.9 & 1.1 / 0.2 & 3.1 / 0.8 & 11.0 / 3.8 & 14.3 / 5.7 \\
    DTR-F1 & Prescription & Precision & \textbf{30.9} / \textbf{17.1} & 20.1 / 13.0 & 1.9 / 0.9 & 2.0 / 0.6 & 17.5 / 10.8 & 23.6 / 14.8 \\
    DTR-F1 & Prescription & Recall & \textbf{29.9} / \textbf{17.4} & 14.5 / 9.5 & 1.5 / 0.7 & 1.1 / 0.3 & 18.3 / 11.7 & 23.2 / 15.4 \\
    DTR-F1 & Prescription & F1 & \textbf{29.8} / \textbf{16.8} & 15.6 / 9.8 & 1.6 / 0.8 & 1.3 / 0.4 & 16.8 / 10.5 & 22.4 / 14.2 \\
    DTR-F1 & Dosage & MAE & 4.2 / 4.3 & 2.7 / 2.5 & 0.4 / 0.2 & \textbf{0.0} / \textbf{0.0} & 3.0 / 2.5 & 3.8 / 3.6 \\
    DTR-F1 & Dosage & Cosine & \textbf{28.7} / \textbf{15.8} & 14.8 / 9.0 & 1.4 / 0.6 & 0.1 / 0.0 & 15.9 / 9.9 & 20.9 / 13.0 \\
    \midrule
    % --- DR
    DR & Syndrome & Accuracy & \textbf{84.2} / 34.5 & 64.6 / 32.0 & 32.5 / 17.3 & 47.2 / 28.5 & 73.8 / 34.3 & 80.7 / \textbf{41.5} \\
    DR & Treatment & Accuracy & \textbf{78.4} / 19.0 & 62.6 / 28.5 & 38.5 / 20.5 & 48.3 / 27.5 & 72.0 / \textbf{32.3} & 72.7 / 26.5 \\
    DR & Prescription & Accuracy & \textbf{87.4} / 43.3 & 36.2 / 25.3 & 18.1 / 15.8 & 23.0 / 18.5 & 65.6 / 39.8 & 80.5 / \textbf{48.0} \\
    \midrule
    \textbf{Average} &  &  & \textbf{47.5} / \textbf{27.2} & 30.3 / 20.7 & 14.6 / 11.2 & 11.7 / 8.6 & 35.3 / 23.7 & 33.5 / 21.5 \\
    \bottomrule
    \end{tabular}
  % }
\end{table*}

\clearpage
\subsection{Difficulty Profiles and Hard-Subset Design}

\begin{figure}[!htbp]
  \centering
  \includegraphics[width=0.99\textwidth]{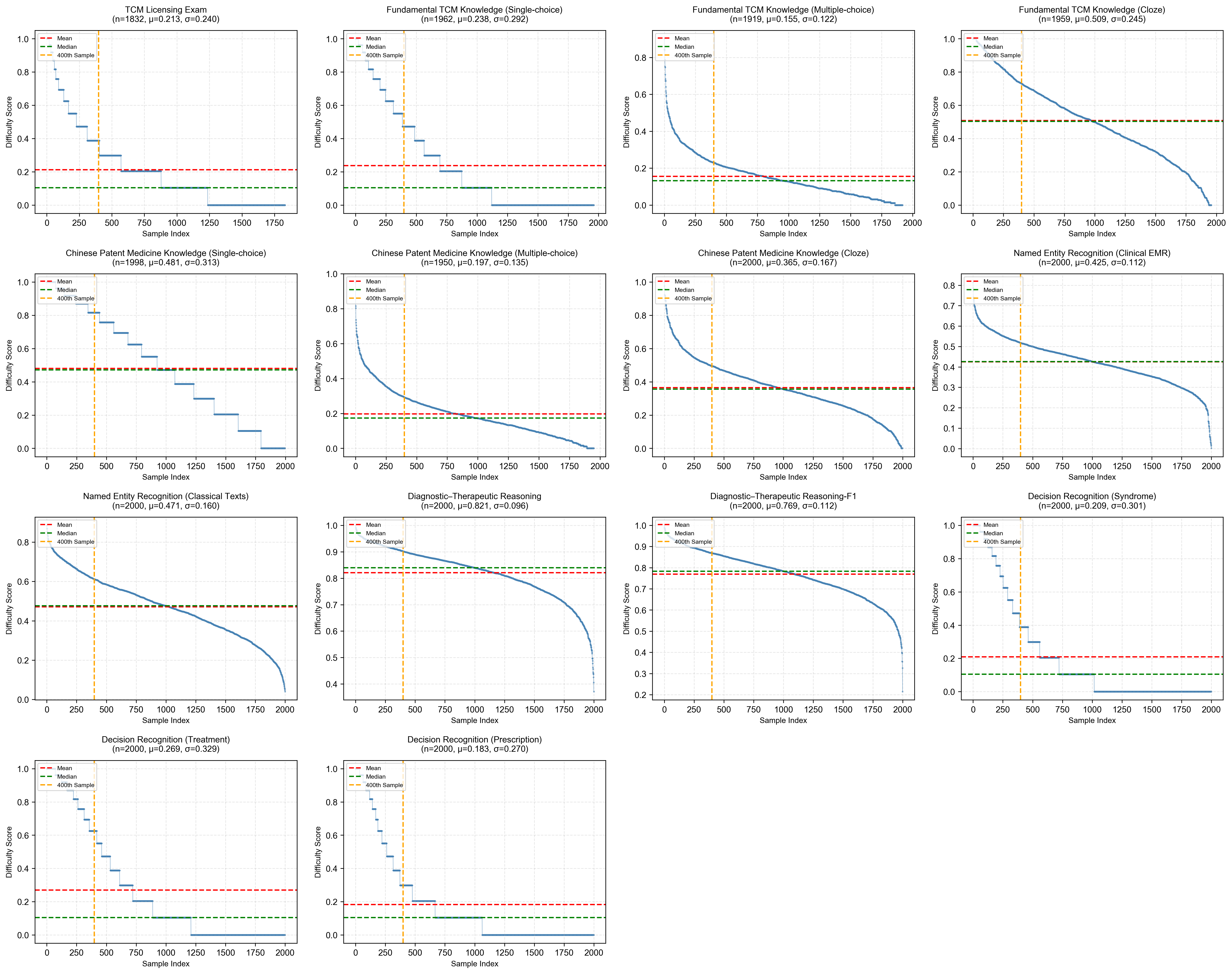}
  \caption{Within each task, items are ranked by decreasing difficulty score (blue curve); the red and green dashed lines show the mean and median, and the orange vertical line marks the top 400 items selected for LingLanMedBench-Hard.}
  \label{fig:difficulty_distribution}
\end{figure}

Figure~\ref{fig:difficulty_distribution} visualizes item difficulty by sorting each dataset’s instances by an empirical difficulty score (blue curve), marking the mean and median with dashed lines, and using the 400th item as the Hard cutoff. Licensing and single-choice knowledge tasks show steep early declines that indicate many easy recall items with a concentrated hard head, whereas multi-choice knowledge exhibits a more gradual slope and broader dispersion. Cloze tasks remain elevated longer before tapering, reflecting lexical sensitivity that keeps many items moderately difficult. In information extraction, classical texts sit consistently higher and decay more slowly than EMRs, revealing a heavier tail due to archaic vocabulary and elliptical syntax. In diagnostic–therapeutic reasoning, synonym-tolerant matching lowers the curve relative to strict scoring yet leaves a persistent hard tail driven by compositional prescription errors and dosage proportionality. Decision-recognition panels display stepped profiles, consistent with items that are either straightforward or distinctly challenging. Collectively, the Hard subsets lie above both mean and median across tasks, isolating the high-difficulty region while preserving each task’s characteristic sources of difficulty.

\clearpage
\subsection{Additional dataset statistics}

\begin{figure}[!htbp]
  \centering
  \includegraphics[width=0.8\textwidth]{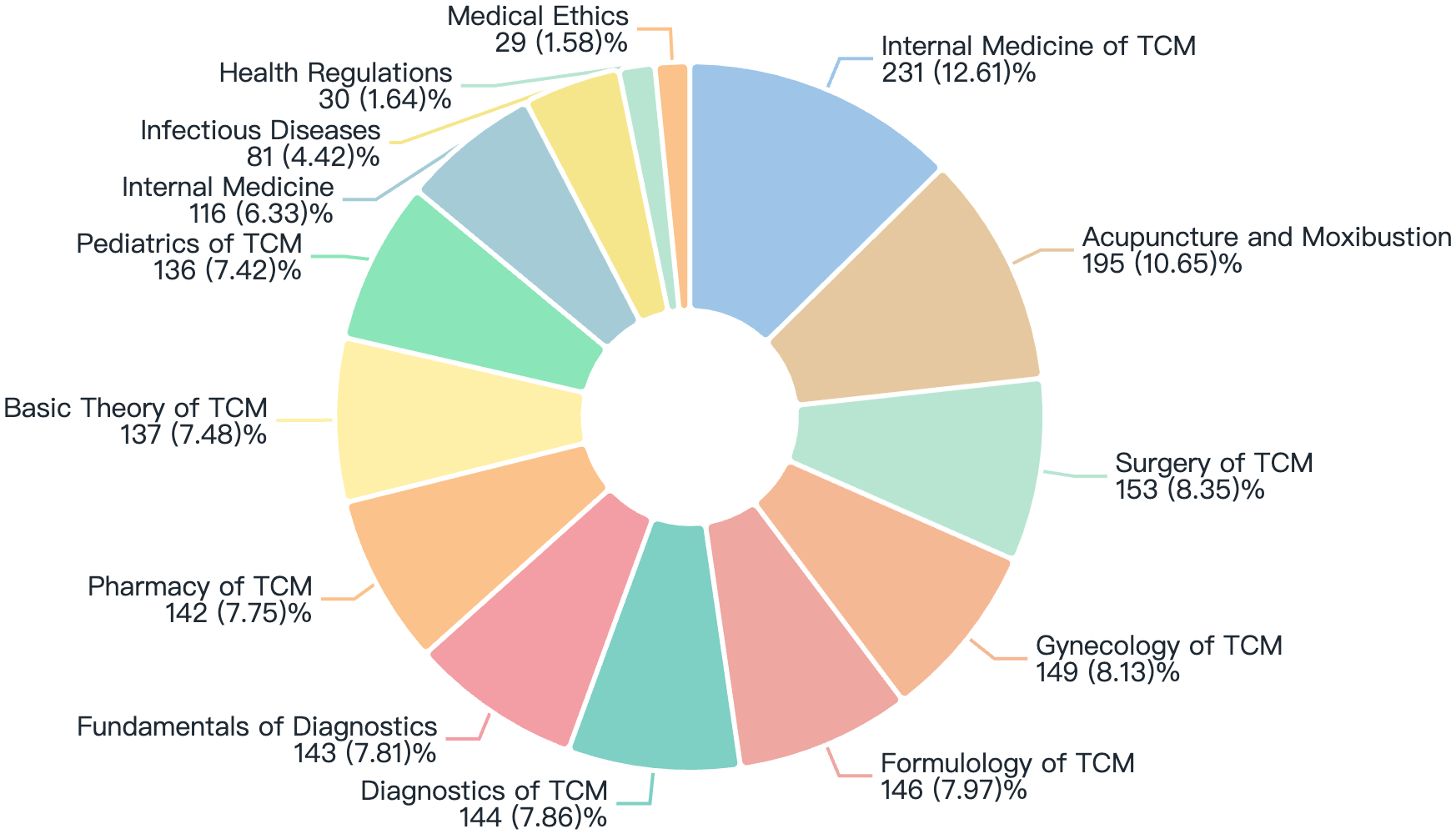}
  \caption{Distribution of items across the 14 subjects in the TLE (TCM Licensing Examination) dataset.}
  \label{fig:tle}
\end{figure}

\begin{figure}[!htbp]
  \centering
  \includegraphics[width=\textwidth]{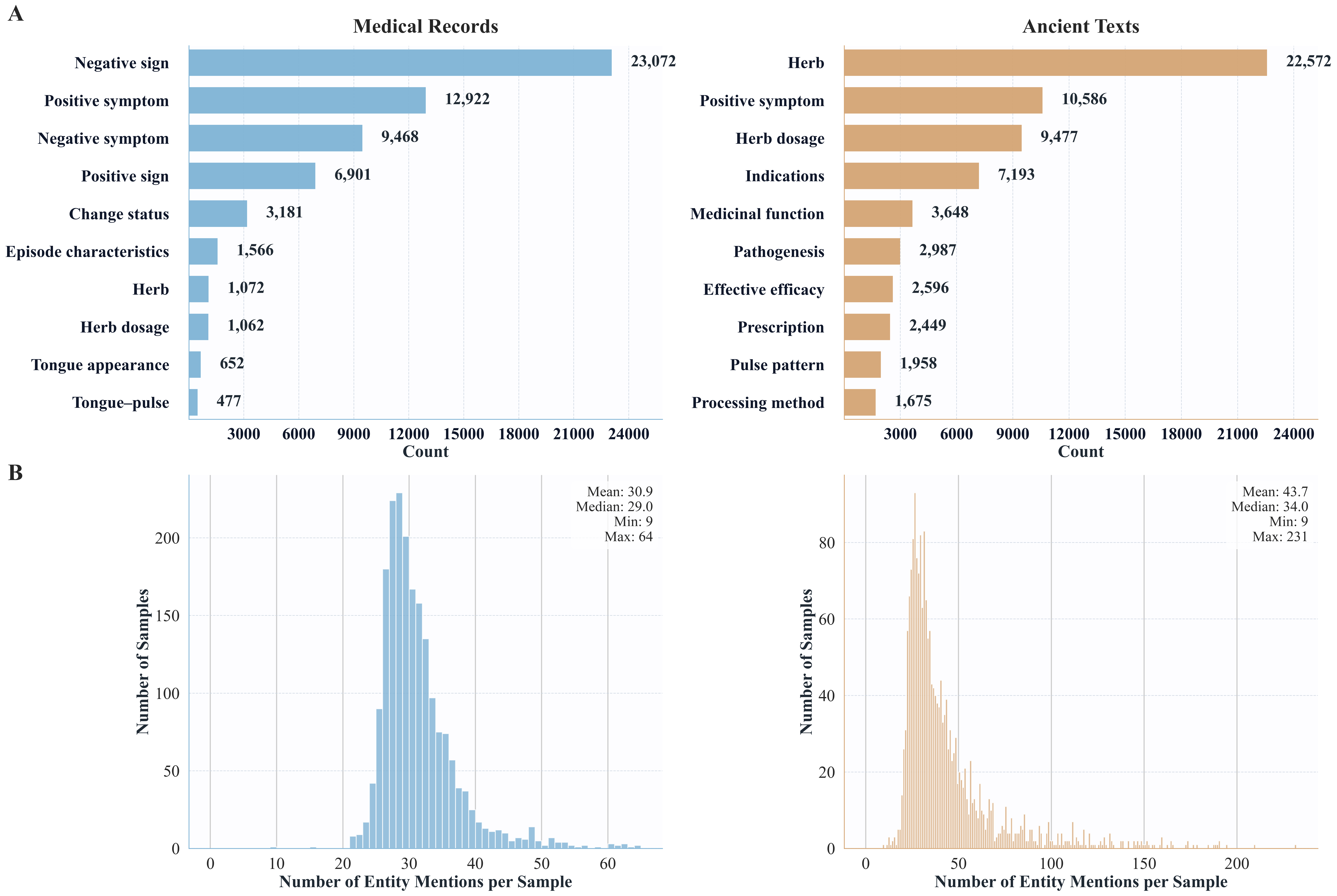}
  \caption{Statistical analysis of entity distribution in IE tasks. (A) Top 10 most frequently mentioned entity types in modern medical records (left, blue) and ancient medical texts (right, orange). Vertical bars show absolute mention counts per category. (B) Distribution of entity mentions per sample for medical records (left, blue) and ancient texts (right, orange).}
  \label{fig:entity_statistics}
\end{figure}

\subsection{Additional descriptions of baseline models}

To establish reliable baselines, we evaluated fourteen representative LLMs spanning diverse architectures, including dense and mixture-of-experts (MoE) models, model scales, and training paradigms, including supervised fine-tuning (SFT), reinforcement learning from human feedback (RLHF), and explicit reasoning training.

\textbf{Qwen3 series.}
Qwen3-4B, Qwen3-8B, Qwen3-14B, Qwen3-32B, Qwen3-30B-A3B, Qwen3-Next-80B-A3B-Thinking, and Qwen3-235B-A22B are members of the Qwen3 series, which includes dense and MoE variants with multilingual pretraining and an optional \emph{thinking} mode for deliberative inference.  
We use official chat checkpoints where available and enable the family’s thinking capability for consistency with other reasoning-enabled baselines.

\textbf{DeepSeek series.}
DeepSeek-R1 is a reasoning-focused model trained via reinforcement learning to elicit multi-step latent reasoning without extensive supervised traces.  
DeepSeek-V3.1-Think is a reasoning-enabled variant based on the DeepSeek MoE architecture, evaluated in its think mode to standardize deliberative decoding across families.

\textbf{Baichuan-M2-32B} is a 32-billion-parameter model developed on Qwen2.5-32B-Base.  
It incorporates a medical augmented-reasoning framework featuring a large-verifier mechanism and Group Relative Policy Optimization (GRPO)-style optimization to enhance stability and factual consistency in complex clinical reasoning tasks.
The model integrates extensive medical knowledge during supervised fine-tuning and reinforcement alignment, aiming to improve interpretability and robustness in diagnostic reasoning.

\textbf{GPT series}
GPT-5 and GPT-5-mini are the latest models in the GPT series.
GPT-OSS-20B and GPT-OSS-120B are open-weight reasoning models built on an efficient MoE architecture.
They are trained via large-scale distillation and reinforcement learning, optimized for agentic capabilities such as research browsing, Python tool use, and function calling.  
Both models employ a structured chat format for robust instruction following and demonstrate strong performance across reasoning, coding, and safety benchmarks.

\textbf{Chinese Medical and TCM LLMs}
Since LingLanMedBench is a newly released benchmark and most of its items had not appeared in public internet corpora before evaluation, it poses a high level of difficulty. Existing Chinese medical and TCM-focused LLMs are generally smaller and less often equipped with dedicated reasoning modes than contemporary frontier models, which may contribute to their performance gap. Therefore, we report a representative subset of these models in Supplementary Table~S1.

\clearpage
\subsection{Detailed data curation and quality assurance}
\label{sec:supp_curation}

Across all domains, LingLanMedBench applied a unified preprocessing pipeline including text normalization, de-duplication, automatic quality screening, and expert verification. 
All clinical materials were de-identified to ensure privacy compliance, and low-quality or ambiguous entries were systematically excluded prior to inclusion.

\subsubsection{Task Definition and Taxonomy}

LingLanMedBench establishes a unified taxonomy encompassing 13 subtasks (as shown in Figure~2 of the main manuscript) across five major domains of TCM, with a total of 44 quantitative evaluation dimensions.
Each subtask is designed to capture distinct dimensions of model capability, ranging from factual recall and symbolic association to structured extraction and clinical reasoning, thereby ensuring a balanced and multi-perspective assessment framework.
All subtasks adopt standardized input–output formats and metric definitions, enabling cross-model comparability and reproducible benchmarking.

\subsubsection{Expert Annotation and Verification}

All datasets in LingLanMedBench underwent a rigorous expert validation process to ensure factual accuracy, terminological precision, and consistency with established principles of TCM. The review framework was designed to eliminate ambiguity, enforce theoretical correctness, and guarantee linguistic clarity.
During annotation and verification, experts were prohibited from consulting LLMs or other automated assistants to avoid bias. Instead, they consulted authoritative references, including the Pharmacopoeia of the People's Republic of China, Differential Diagnosis of TCM Symptoms, and national TCM Teaching Materials.
The review guidelines emphasized semantic precision and task-specific rigor. To ensure question clarity, all stems were revised to remove ambiguity and to align precisely with the intended evaluation target; for example, vague fill-in-the-blank items were reformulated to specify whether the expected answer is a syndrome or a symptom. Terminological accuracy was enforced by correcting misuse of TCM terms (such as conflating syndromes with diseases or symptoms) to canonical phrasing. For answer validation, reference answers were verified against standard textbooks and authoritative sources; when multiple answers were acceptable, they were listed with enumeration marks to preserve complete semantic coverage. For option revision, ambiguous or partially correct multiple-choice options were replaced with unambiguously correct or incorrect alternatives to enhance discriminative validity. As part of quality control, items with irreparable logical or semantic flaws were excluded from the final dataset.
This standardized protocol ensures that LingLanMedBench items are theoretically sound, linguistically precise, and pedagogically robust, thereby minimizing uncertainty and maximizing reproducibility across diverse LLM evaluation settings.

\subsubsection{Knowledge-Oriented Tasks}
This subsection describes the curation and quality assurance for the three knowledge-oriented suites, including the TCM Licensing Examination (TLE), Fundamental TCM Knowledge (FTK), and Chinese Patent Medicine Knowledge (CPMK).

For TLE, all items were standardized into a single-choice format with a unique gold answer and a normalized subject tag. We removed duplicates through fingerprinting of stems and explanations, corrected typography and punctuation, and rewrote unclear stems to eliminate answer ambiguity. Option sets were audited to ensure one and only one valid key, and distractors that were partially correct or definition-overlapping were replaced. Items that could not be unambiguously repaired were discarded, yielding a final pool of 1{,}832 high-quality questions.
The corpus spans 14 subjects (Supplementary Fig.~S2).

For the FTK task, we synthesize questions from an in-house TCM knowledge graph by systematically traversing its triples. 
For each head entity, we retrieve its associated triples in fixed-size batches of 20, when an entity is linked to more than 20 triples, we iterate over successive, non-overlapping batches of size 20, treating each batch as an independent background context. 
Using task-specific prompts and the Qwen3-32B model, we generate a large pool of candidate items spanning single-choice, multiple-choice, and cloze (fill-in-the-blank) formats, totaling several hundred thousand entries.
For the CPMK task, we start from a curated knowledge base of Chinese patent medicines and use, for each medicine, its associated structured knowledge as the background context. We then apply the same synthesis pipeline as in FTK to produce hundreds of thousands of candidate items that cover clinical usage, contraindications, administration, and therapeutic indications.
All candidates undergo machine screening through a combination of regex-based heuristics (e.g., proportion of Latin characters, length constraints) and LLM-based quality scoring to remove redundant or low-quality items.
Licensed TCM practitioners then review the remaining items to ensure linguistic clarity, terminological accuracy, and clinical plausibility. 
After multi-stage filtering and verification, 5,840 high-quality FTK items and 5,948 high-quality CPMK items are retained for inclusion in the benchmark.

\subsubsection{Information Extraction Tasks}

The information extraction task comprises two span-annotated corpora drawn from distinct text domains: de-identified modern Chinese electronic medical records (EMR) and classical TCM literature written in Literary Chinese. 
Following a human–machine workflow that integrated automatic NER suggestions with expert adjudication, we retained 2,000 samples for each corpus, yielding a total of 4,000 samples.
The suite covers more than one hundred entity types, including positive and negative symptoms, physical signs, syndromes, Chinese herbs, formulas, condition changes, and episode characteristics. The clinical corpus contains 67 entity types with 11,886 mentions; the classical corpus contains 56 types with 19,876 mentions; in total there are 106 unique types. As shown in Supplementary Fig.~S3A, symptoms and physical signs are most frequent in EMR, reflecting the emphasis on symptom/sign documentation in clinical records; in the classical corpus, herbs and symptoms dominate, consistent with the prevalence of casebooks authored by renowned physicians. Supplementary Fig.~S3B presents the per-sample mention distribution: most samples contain 30–50 entities, and the densest classical case includes up to 231 entity mentions.

\subsubsection{Diagnostic and Therapeutic Decision-Making Tasks}

The auxiliary clinical decision component comprises two tasks: Diagnostic and Therapeutic Reasoning (DTR) and Decision Recognition (DR). 
From an initial pool of over 13,000 case records, we retained those with detailed documentation, specifically complete four-diagnoses notes and prescriptions containing at least five herbs, resulting in a curated DTR set of 2,000 high-quality cases.
This set covers 1,562 distinct syndromes, 1,811 treatment principles, and 1,631 Chinese herbs.
DTR integrates four interdependent subtasks that trace the clinical workflow: syndrome differentiation, treatment principle selection, prescription recommendation, and dosage estimation. Taken together, these subtasks provide a comprehensive assessment of model competence in integrative clinical reasoning and actionable decision-making.
Given that free-form prediction in DTR must search over extremely large label spaces conditioned only on the case text, we also construct DR by reframing each clinical problem as a single-choice question to measure cognitive alignment with expert reasoning while minimizing confounds from open-ended decoding. For DR, each case yields three items (syndrome, treatment principle, prescription), each with one correct option and four distractors, and is evaluated by accuracy. Option sets are built via embedding-based retrieval: we encode the query case with Qwen3-0.6B-Embedding, compute cosine similarity over a large indexed corpus, and collect the top 1,000 nearest neighbors. The correct option is the ground-truth label of the query case; distractors are sampled from labels observed among the neighbors with lower semantic similarity to the query, ensuring they remain contextually plausible yet discriminative.

In total, LingLanMedBench provides a comprehensive evaluation matrix integrating 44 indicators across 13 subtasks.
This taxonomy ensures that every dimension of TCM knowledge and reasoning, across the spectrum from classical textual comprehension to modern clinical decision-making, is quantitatively represented.

\subsubsection{Hard subset construction}

To evaluate model performance under more challenging conditions, we construct a high-difficulty subset named LingLanMedBench-Hard. Item difficulty is estimated from per-instance scores across 14 models. For each item, we compute the across-model mean \(\mu\) and variance \(\sigma\), and define a composite difficulty score \(D=(1-\mu)+\lambda\sigma\), where \(\lambda=0.5\). Items are then ranked within each task in descending order of \(D\), and the top 400 highest-difficulty items from each task are selected. In total, LingLanMedBench-Hard covers 13 tasks and contains 5,200 samples. The overall difficulty distribution across tasks is illustrated in Supplementary Fig.~S1.

\clearpage
\subsection{Detailed evaluation metrics}
\label{sec:supp_metrics}

\setcounter{equation}{0}
\renewcommand{\theequation}{S\arabic{equation}}
\renewcommand{\theHequation}{supp.\arabic{equation}}

LingLanMedBench employs a unified metric matrix to ensure consistent, quantitative, and interpretable evaluation across heterogeneous task types.  
Each metric is aligned with the structural form of its corresponding task, allowing for fair comparison among classification, structured prediction, and regression subtasks.

\subsubsection{Accuracy for Single- and Multiple-Choice}
Accuracy is applied to discrete categorical tasks, including the TLE, FTK, CPMK, and DR subtasks.
For single-choice questions, each instance has a unique correct option; for multiple-choice questions, an answer is considered correct only when all correct options are selected and no incorrect options are chosen:

\begin{equation}
\mathrm{Accuracy} = \frac{N_{\mathrm{correct}}}{N_{\mathrm{total}}}
\end{equation}

where $N_{\mathrm{correct}}$ and $N_{\mathrm{total}}$ denote the number of correctly answered and total questions, respectively.

\subsubsection{Precision, Recall, and F1 for Cloze, Multiple-Choice IE, DTR, and DTR-F1}

For any instance, let $TP_i$, $FP_i$, and $FN_i$ denote true, false-positive, and false-negative counts as defined by the task-specific matching rule below. Precision, recall, and F1 are

\begin{equation}
\mathrm{Precision}_i=\frac{TP_i}{TP_i+FP_i},\qquad
\mathrm{Recall}_i=\frac{TP_i}{TP_i+FN_i},\qquad
\mathrm{F1}_i=\frac{2\,\mathrm{Precision}_i\,\mathrm{Recall}_i}{\mathrm{Precision}_i+\mathrm{Recall}_i}
\end{equation}

and the dataset score is the macro average $\frac{1}{N}\sum_{i=1}^{N}\mathrm{F1}_i$.

\textbf{(1) Cloze (character-level matching).}
Given a predicted string $\hat{s}$ and gold string $s$, treat them as multisets of characters. For each character $c$,
let $n_{\hat{s}}(c)$ and $n_{s}(c)$ be their multiplicities. Define

\begin{equation}
TP_{\text{char}}=\sum_{c}\min\big(n_{\hat{s}}(c),n_{s}(c)\big),
FP_{\text{char}}=\sum_{c}\max\big(0,n_{\hat{s}}(c)-n_{s}(c)\big),
FN_{\text{char}}=\sum_{c}\max\big(0,n_{s}(c)-n_{\hat{s}}(c)\big)
\end{equation}

Use the unified formulas above with $(TP_i,FP_i,FN_i)=(TP_{\text{char}},FP_{\text{char}},FN_{\text{char}})$.
This captures partial correctness and orthographic variation at character granularity.

\textbf{(2) Multiple-Choice.}
Each instance consists of selecting options from a predefined set. Let $Y_i$ denote the gold set of correct options and $\hat{Y}_i$ the predicted set. Define
\begin{equation}
TP_i=|\hat{Y}_i\cap Y_i|,\qquad
FP_i=|\hat{Y}_i\setminus Y_i|,\qquad
FN_i=|Y_i\setminus \hat{Y}_i|.
\end{equation}
Apply the unified formulas. This regime measures exact set agreement between predicted and correct options.

\textbf{(3) Information Extraction (instance-level multiset matching).}
Each instance is a multiset of typed surface forms over the universe $U$ of pairs $u=(\text{type},\text{text})$.
Let $G_i$ and $\hat{G}_i$ be gold and predicted multisets with multiplicities $c_i(u)$ and $\hat{c}_i(u)$. Define
\begin{equation}
TP_i=\sum_{u\in U}\min\!\big(c_i(u),\hat{c}_i(u)\big),\quad
FP_i=\sum_{u\in U}\max\!\big(0,\hat{c}_i(u)-c_i(u)\big),\quad
FN_i=\sum_{u\in U}\max\!\big(0,c_i(u)-\hat{c}_i(u)\big)
\end{equation}
where matching requires exact equality of entity type and normalized surface form. Apply the unified formulas.

\textbf{(4) DTR.}
For syndrome differentiation, treatment principle, and prescription recommendation, each instance is a set of canonical labels without duplicates (order ignored).
After schema-preserving normalization, counts are computed with a one-to-one matching rule that treats two labels as the same if either string contains the other.
With gold set $Y_i$ and predicted set $\hat{Y}_i$ after schema-preserving normalization,
\begin{equation}
TP_i=|\hat{Y}_i\cap Y_i|,\qquad
FP_i=|\hat{Y}_i\setminus Y_i|,\qquad
FN_i=|Y_i\setminus \hat{Y}_i|.
\end{equation}
Apply the unified formulas. This regime measures compositional correctness under label agreement.

\textbf{(5) DTR-F1 (synonym-tolerant matching).}
To discount superficial lexical variation, align the predicted and gold \emph{sets} by one-to-one bipartite matching at the instance level. 
For any predicted label $\hat{y}$ and gold label $y$, compute the character-level F1 $\mathrm{F1}_{\text{char}}(\hat{y},y)$ as in the cloze definition. 
Construct a bipartite graph with edges where $\mathrm{F1}_{\text{char}}(\hat{y},y)\ge\tau$ (threshold $\tau$ = 0.7) and take a maximum-cardinality matching $M_i$. Set
\begin{equation}
TP_i=|M_i|,\qquad FP_i=|\hat{Y}_i|-|M_i|,\qquad FN_i=|Y_i|-|M_i|,
\end{equation}
then use the unified formulas. This preserves non-duplicated sets and enforces one-to-one alignment while tolerating near-synonymous surface forms.

\subsubsection{Dosage Estimation}

The dosage estimation task assesses the quantitative and proportional consistency between predicted and reference prescriptions.  
Given the vast combinatorial space of herbal prescriptions in TCM, achieving exact herb-level correspondence remains highly challenging even for advanced LLMs.
To mitigate this difficulty, we introduce flexible alignment strategies by employing both inclusion-based matching (as used in DTR) and character-level F1 matching (as used in DTR-F1), thus ensuring that dosage evaluation remains fair and robust when minor lexical variations occur in herb names.

Each predicted herb is aligned with at most one reference herb according to the following criteria:

(1) Inclusion-based matching: if the predicted herb name is contained within the reference herb name, or vice versa (e.g., ``Ginseng'' and ``Ginseng Root''), the prediction is considered correct.

(2) Character-level F1 matching: if inclusion does not hold, we compute the character-level F1 between the predicted and reference herb names as in the DTR-F1 evaluation; if it exceeds a threshold $\tau=0.7$, the herb is also considered correct.
This matching mechanism reduces the impact of superficial naming variation while maintaining one-to-one alignment between herbs, thus enabling consistent dosage evaluation.

Based on this aligned herb mapping, quantitative metrics are then computed to assess both absolute and proportional accuracy of dosage prediction.

(1) Mean Absolute Error (MAE).
After alignment, the Mean Absolute Error quantifies the average deviation in dosage between matched herb pairs:
\begin{equation}
\mathrm{MAE}_i =
\begin{cases}
\dfrac{1}{K_i}\sum_{k=1}^{K_i} \big| \hat{d}_{ik} - d_{ik} \big|, & K_i > 0, \\[6pt]
0, & \text{otherwise,}
\end{cases}
\end{equation}
where $K_i$ denotes the number of matched herbs in instance $i$, and $\hat{d}_{ik}$ and $d_{ik}$ are predicted and reference dosages, respectively.  
MAE directly measures absolute quantitative accuracy. It provides a precise reflection of dosage deviation when most herbs are correctly matched between prediction and reference.

(2) Cosine Similarity.
To better capture proportional dosage alignment, particularly when partial overlap exists between predicted and reference prescriptions, cosine similarity is also computed.  
Aligned dosage vectors $\hat{\mathbf{d}}_i$ and $\mathbf{d}_i$ are constructed using inclusion- or character-F1-based matching; unmatched herbs are padded with zeros.  
For each instance:
\begin{equation}
\mathrm{Cos}_i(\hat{\mathbf{d}}_i, \mathbf{d}_i) =
\begin{cases}
\dfrac{\hat{\mathbf{d}}_i^\top \mathbf{d}_i}{\lVert \hat{\mathbf{d}}_i\rVert_2 \, \lVert \mathbf{d}_i\rVert_2}, &
\lVert \hat{\mathbf{d}}_i\rVert_2 > 0 \ \wedge\ \lVert \mathbf{d}_i\rVert_2 > 0,\\[6pt]
1, & \lVert \hat{\mathbf{d}}_i\rVert_2 = 0 \ \wedge\ \lVert \mathbf{d}_i\rVert_2 = 0,\\[6pt]
0, & \text{otherwise.}
\end{cases}
\end{equation}
Cosine similarity focuses on relative dosage proportions rather than absolute quantities.  
It remains informative even when some herbs are missing or mismatched, providing a complementary perspective on the model’s dosage reasoning ability.

Together, MAE and cosine similarity jointly characterize both absolute and proportional dimensions of dosage prediction, ensuring a comprehensive evaluation of quantitative reasoning in prescription generation.

% \end{document}

\end{document}